\documentclass{article} 
\usepackage{iclr2025_conference,times}

\usepackage{amsmath,amsfonts,bm}

\def\eqref#1{equation~\ref{#1}}

\def\1{\bm{1}}

\DeclareMathAlphabet{\mathsfit}{\encodingdefault}{\sfdefault}{m}{sl}
\SetMathAlphabet{\mathsfit}{bold}{\encodingdefault}{\sfdefault}{bx}{n}

\usepackage{hyperref}
\usepackage{url}
\usepackage{booktabs}
\usepackage{graphicx}

\usepackage{url}
\usepackage[flushleft]{threeparttable}
\usepackage{times}
\usepackage{latexsym}
\usepackage{multirow}
\usepackage{graphicx}  
\usepackage{pgfplots}
\pgfplotsset{compat=1.12}
\usepackage{amsmath}
\usepackage{multicol}
\usepackage{amssymb}
\usepackage{booktabs}
\usepackage{xspace}
\usepackage{wrapfig}
\usepackage{arydshln}
\usepackage{tabularray}
\usepackage{subcaption}
\usepackage{color}
\usepackage{colortbl,array,xcolor}

\usepackage{arydshln}
\usepackage{amsmath}
\usepackage{enumitem}
\usepackage{amssymb}
\usepackage{xspace}
\usepackage{mathtools}
\usepackage{amsthm}
\usepackage{caption}
\usepackage{subcaption}
\usepackage{algorithm}
\usepackage{algorithmic}
\usepackage{listings}
\usepackage{multirow}
\usepackage{multicol}
\usepackage[normalem]{ulem}
\usepackage{mdframed}
\usepackage{subfiles}
\usepackage{soul}
\usepackage{wrapfig}
\usepackage{titletoc}
\usepackage{tikz}
\usepackage{booktabs}

\definecolor{gred}{HTML}{cc0200}
\definecolor{ggreen}{HTML}{4C9F26}
\definecolor{Gray}{gray}{0.90}
\newcommand\cg{\cellcolor{Gray}}

\newcolumntype{a}{>{\columncolor{Gray}}c}

\title{ADePT: Adaptive Decomposed Prompt Tuning for Parameter-Efficient Fine-tuning}

\author{Pengwei Tang, Xiaolin Hu, Yong Liu\thanks{Corresponding Author.}  \\
Renmin University of China, Beijing, China\\
\texttt{\{tangpwei,xiaolinhu,liuyonggsai\}@ruc.edu.cn} \\
}

\iclrfinalcopy 
\begin{document}

\maketitle

\begin{abstract}

Prompt Tuning (PT) enables the adaptation of Pre-trained Large Language Models (PLMs) to downstream tasks by optimizing a small amount of soft virtual tokens, which are prepended to the input token embeddings. Recently, Decomposed Prompt Tuning (DePT) has demonstrated superior adaptation capabilities by decomposing the soft prompt into a shorter soft prompt and a pair of low-rank matrices. The product of the pair of low-rank matrices is added to the input token embeddings to offset them. Additionally, DePT achieves faster inference compared to PT due to the shorter soft prompt. However, in this paper, we find that the position-based token embedding offsets of DePT restrict its ability to generalize across diverse model inputs, and that the shared embedding offsets across many token embeddings result in sub-optimization. To tackle these issues, we introduce \textbf{A}daptive \textbf{De}composed \textbf{P}rompt \textbf{T}uning (ADePT), which is composed of a short soft prompt and a shallow token-shared feed-forward neural network. ADePT utilizes the token-shared feed-forward neural network to learn the embedding offsets for each token, enabling adaptive embedding offsets that vary according to the model input and better optimization of token embedding offsets. This enables ADePT to achieve superior adaptation performance without requiring more inference time or additional trainable parameters compared to vanilla PT and its variants. In comprehensive experiments across 23 natural language processing tasks and 4 typical PLMs of different scales, ADePT consistently surpasses the other leading parameter-efficient fine-tuning methods, and even outperforms the full fine-tuning in certain scenarios. We also provide a theoretical analysis towards ADePT. Code is available at \url{https://github.com/HungerPWAY/ADePT}. 
\end{abstract}

\section{Introduction}
\label{sec:intro}
Recently, Pre-trained Large Language Models (PLMs)~\citep{t5,touvron2023llama} have seen rapid development, with commonly used models now on the scale of hundreds of millions and billions of parameters. Full fine-tuning (FT) of these PLMs requires substantial GPU resources, which is a common challenge faced by both academia and industry. To alleviate this resource-intensive issue, Parameter-Efficient Fine-Tuning (PEFT) \citep{houlsby2019parameter,NEURIPS2022_0cde695b,hu2021lora,zaken2021bitfit} methods have gained significant attention and have seen breakthrough progress. These PEFT methods tune only a small amount of the internal parameters of a model or extra parameters, allowing PLMs to adapt effectively to target downstream tasks while maintaining performance comparable to FT. 

\begin{figure}[!t]
\centering
\captionsetup[subfigure]{font=scriptsize}
    \begin{subfigure}[b]{0.31\textwidth}
        \centering
  \raisebox{6ex}{\includegraphics[width=0.89\textwidth]{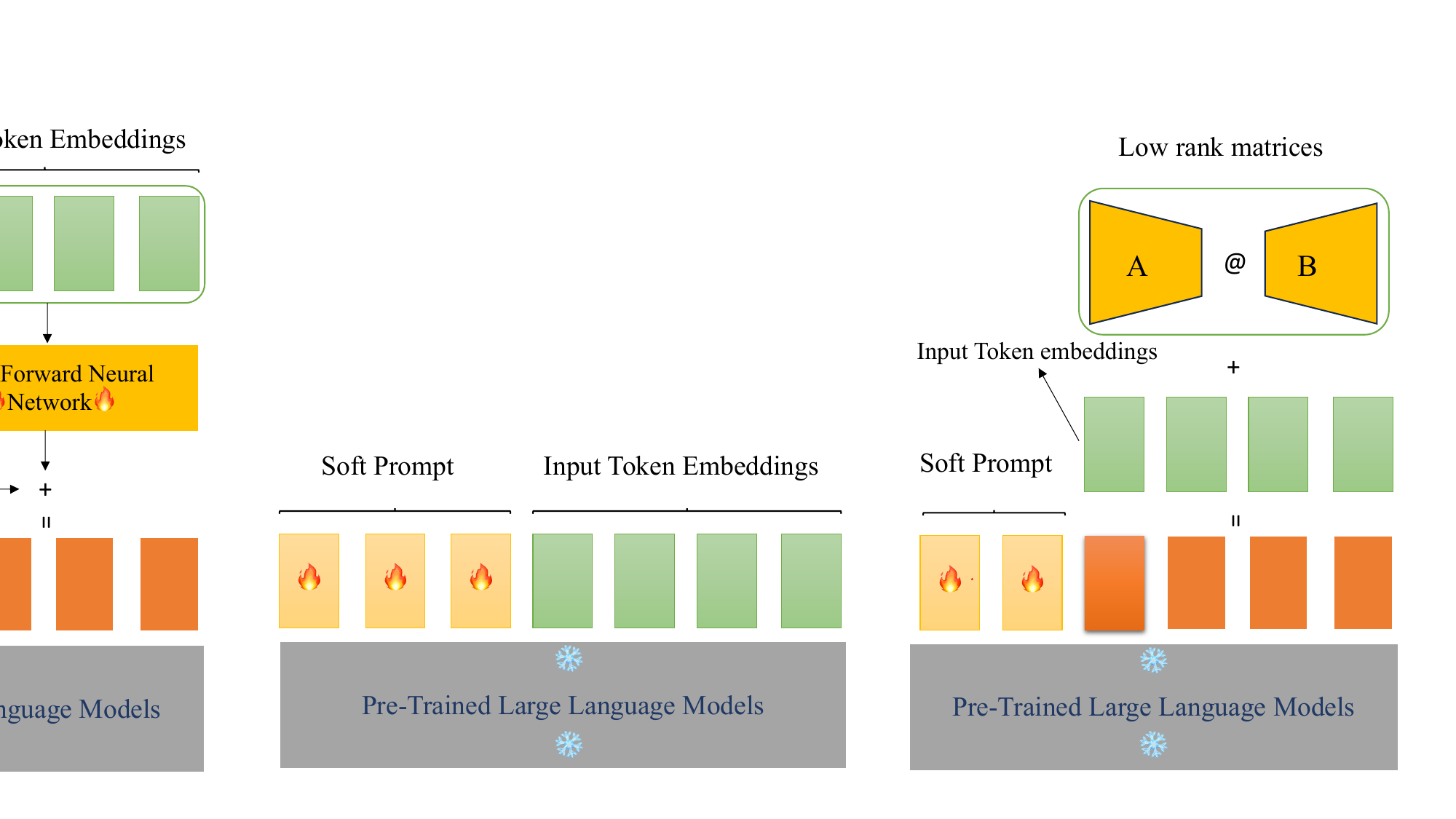}}
        \caption{\textbf{Prompt Tuning}}
        \label{fig:PT}
    \end{subfigure}
    \hfill
    \begin{subfigure}[b]{0.31\textwidth}
        \centering
        \includegraphics[width=0.88\textwidth]{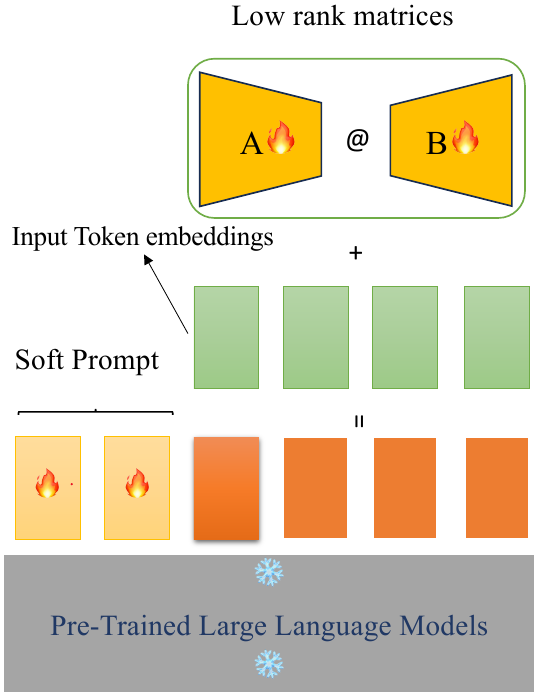}
        \caption{\textbf{Decomposed Prompt Tuning}}
        \label{fig:DePT}
    \end{subfigure}
    \hfill
    \begin{subfigure}[b]{0.31\textwidth}
        \centering
        \includegraphics[width=0.89\textwidth]{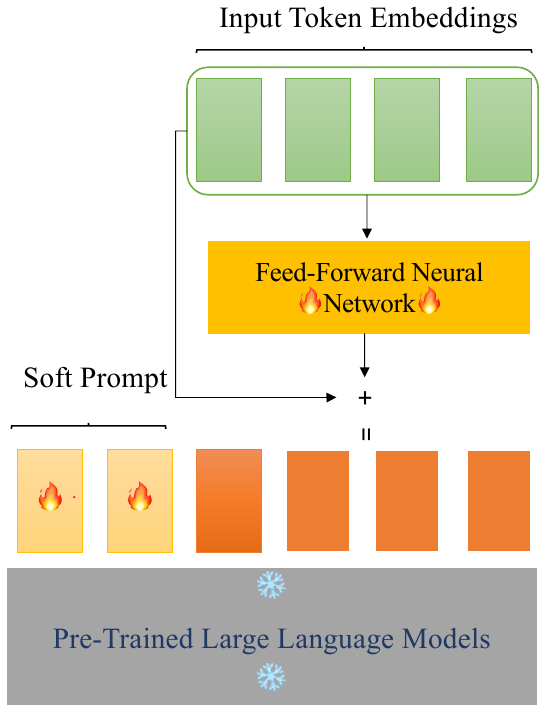}
        \caption{\textbf{Adaptive Decomposed Prompt Tuning}}
        \label{fig:ADePT}
    \end{subfigure}
    \caption{The overview of Prompt Tuning (PT), Decomposed Prompt Tuning, and Adaptive Decomposed Prompt Tuning (ADePT). PT uses a soft prompt prepended to input token embeddings. DePT uses a short soft prompt and offsets the input token embeddings using a pair of low-rank matrices. ADePT uses a short soft prompt and offsets the input token embedding using a token-shared shallow feed-forward neural network. ADePT can adaptively give input token embedding offsets based on input tokens, while DePT can only give position-based input token embedding offsets. Moreover, the use of a short soft prompt makes DePT and ADePT faster during inference.}
    \label{fig:overview}
    \vspace{-1.5em}
\end{figure}

The vanilla Prompt Tuning (PT) \citep{lester-etal-2021-power} uses a trainable soft prompt prepended to the input token embeddings \citep{lester-etal-2021-power}, as shown in Figure \ref{fig:PT}. 
The few trainable parameters make PT one of the mainstream methods for parameter-efficient fine-tuning. The improvements to PT can be categorized into four paths: the first path involves adding soft prompts to each layer of one PLM \citep{li-liang-2021-prefix}; the second path involves stabilizing the optimization of soft prompt through a shallow network with a residual connection \citep{ResidualPrompt}; the third path involves using soft prompts that had already been trained by other methods for transfer learning \citep{vu-etal-2022-spot,asai-etal-2022-attempt,MPT}; the fourth path uses input token embedding offsets to map the input token embedding into better embedding space \citep{shidept}. The first three approaches either require increasing trainable parameters or necessitate additional transfer learning, while the fourth approach requires neither. 
For the fourth path, Decomposed Prompt Tuning (DePT) \citep{shidept} pioneers the shift in attention from soft prompt to applying input token embeddings offsets to the input token embeddings. As shown in Figure \ref{fig:DePT}, 
DePT learns embedding offsets $\Delta \bm{ E} = \bm{A} \bm{B} = [\Delta \bm{e}_1, \Delta \bm{e}_2, \cdots, \Delta \bm{e}_s]$ by optimizing a pair of low-rank matrices $\bm{A} \in \mathbb{R}^{s \times r_s}$ and $\bm{B} \in \mathbb{R}^{r_s \times d}$, where $s$ denotes the length of input tokens, $d$ denotes the dimension of input tokens and $r_s$ denotes the maximum rank of matrices $\bm A$ and $\bm B$. Given the input token embeddings $\bm{E} = [\bm{e}_1, \bm{e}_2, \cdots, \bm{e}_s] \in \mathbb{R}^{s \times d}$, the updated token embeddings $\bm{E}'$ are:
\begin{equation*}
    \bm{E}' = \bm{E} + \Delta \bm{E} = \bm{E} + \bm{A}\bm{B}
\end{equation*}
DePT also learns a short soft prompt $\bm{P}_s \in \mathbb{R}^{l_s \times d}$ with length $l_s$. The input token embeddings $\bm{E}$, after DePT processing, are formulated as $[\bm{P}_s, \bm{E} + \bm{A}\bm{B}]$. In this paper, we focus on improving the fourth path by providing better offsets for the input token embeddings. Additionally, our improvement of the fourth path is orthogonal to the first three and can be integrated with their methods. 

The main contribution of DePT is the use of learnable token embedding offsets, which map the input token embeddings into a more suitable embedding space. DePT has demonstrated promising results across various tasks and PLMs, but it still has limitations. 
The updated token embeddings are formulated as $\bm{E} + \Delta \bm{E} = [\bm{e}_1 + \Delta \bm{e}_1, \bm{e}_2 + \Delta \bm{e}_2, \cdots, \bm{e}_s + \Delta \bm{e}_s]$. We can observe that in a target downstream task, the input token embeddings $\bm{E}$ vary, while the offsets $\Delta \bm{E}$ are position-based and fixed after fine-tuning. For a token $\bm{e}$, its position in the task can be arbitrary, meaning that both $\bm{e} + \Delta \bm{e}_i$ and $\bm{e} + \Delta \bm{e}_j$ can be its updated embedding in this task, where $1 \le i \le s, 1\le j \le s, i \neq j$. This means that within the same task, the same token may have multiple different token embeddings, implying that the token embeddings of DePT violate the uniqueness of token embeddings. In Section \ref{sec:limitations}, we design two experiments that reveal two limitations of DePT: (1) the offsets obtained by tokens at different positions may lead to different prediction outcomes; (2) to ensure that token embeddings are as unique as possible, the offsets in DePT are often much smaller than the input token embeddings, leading to sub-optimization.

Based on the above analysis, the embedding offset for each token should be the same at any position. Additionally, to enhance expressiveness, the offset corresponding to each token should be as unique as possible. Surprisingly, we find that using a function of the token embedding can meet the above requirements, \textit{i.e.}, using the function $f(\bm{e})$ to be the offsets of input token embedding $\bm {e}$. To avoid increasing the number of trainable parameters compared to vanilla PT, we utilize a shallow and narrow feed-forward neural network to approximate the function for offset prediction. As shown in Figure \ref{fig:ADePT}, we propose \textbf{A}daptive \textbf{De}composed \textbf{P}rompt \textbf{T}uning (ADePT), which uses a short soft prompt and a two-layer feed-forward neural network, where the short soft prompt is prepended to the input token embeddings and the feed-forward neural network is shared by all tokens and produces the offset of each token. The use of a short prompt makes ADePT achieve faster inference speeds than the vanilla PT and comparable inference speeds compared to DePT.

In summary, the main contributions of this paper are as follows:
\begin{itemize}
    \item 
    We point out that the limitations of position-based token embedding offsets in DePT.To tackle these issues, we propose Adaptive Decomposed Prompt Tuning (ADePT), which can produce unique token embedding offset for each token. 
    \item ADePT employs token-shared feed-forward neural networks to learn a unique offset for each token, which can adaptively adjust the token embedding offsets based on model input and allow the training parameters to achieve better optimization.
    \item Extensive evaluations across 23 NLP tasks and 4 PLMs of different scales show that ADePT surpasses leading PEFT methods, including full fine-tuning in certain scenarios. 
    We theoretically analyze that ADePT possesses better expressive power than PT and DePT.
\end{itemize}

\section{Related Works}
\textbf{Parameter-Efficient Fine-tuning:} 
PEFT methods adapt PLMs to downstream tasks by optimizing a small number of parameters, significantly reducing computational resource demands. PEFT methods can be divided into four types:(1) Adapter and its variants \citep{houlsby2019parameter,he2022hyperprompt,ruckle-etal-2021-adapterdrop,ivison-peters-2022-hyperdecoders}; (2) Low-rank Adaptation (LoRA) and its variants \citep{hu2021lora, NEURIPS2022_0cde695b,kopiczko2024vera}; (3) Prompt Tuning (PT) and its variants \citep{lester-etal-2021-power,li-liang-2021-prefix,vu-etal-2022-spot,asai-etal-2022-attempt, MPT,ma-etal-2022-xprompt,xiao-etal-2023-decomposed, ResidualPrompt,shidept}; (4) other methods \citep{zaken2021bitfit,guo2021diff}. The adapter inserts new modules into the transformer blocks \citep{houlsby2019parameter}. Hyperformer \citep{he2022hyperprompt} uses a shared hypernetwork to generate task-conditioned adapters, reducing the trainable parameters in multi-task scenarios. AdapterDrop \citep{ruckle-etal-2021-adapterdrop} removes adapters from lower transformer layers during training and inference, which reduces the computation overhead. Hyperdecoders \citep{ivison-peters-2022-hyperdecoders} generate input-specific adapters using a shared decoder for multi-task scenarios. LoRA \citep{houlsby2019parameter} adopts the matrix product of a pair of low-rank matrices to approximate the updates of corresponding parameters. $(\text{IA})^3$ \citep{NEURIPS2022_0cde695b} rescales internal activations only using learned vectors injected into the attention and feedforward modules. Prompt tuning (PT) \citep{lester-etal-2021-power}adapts PLMs to the new downstream tasks by optimizing learnable virtual tokens. LST \citep{sunglst} reduces memory requirements by training a small side network, achieving adaptation without back-propagating through the main network. BitFit \citep{zaken2021bitfit} only tunes the bias of PLMs, which extremely reduces the trainable parameters. Diff pruning learns a sparse ``diff'' vector to modify a small percentage of pre-trained parameters, enabling efficient adaptation \citep{guo2021diff}. Our proposed ADePT is an improved variant of PT, which can also be unified with other PEFT methods.

\textbf{Prompt tuning and its variants:} PT \citep{lester-etal-2021-power} enables the adaptation of PLMs to downstream tasks by learning a soft prompt appended in front of model input.  Prefix-tuning \citep{li-liang-2021-prefix} can be viewed as an extension of Prompt Tuning applied across the entire depth of the model. SPoT \citep{vu-etal-2022-spot} utilizes trained prompts from source tasks as initialization, but it requires extensive search to find the optimal initialization. ATTEMPT \citep{asai-etal-2022-attempt} adopts an attention module to compose the trained prompts of source tasks. MPT \citep{MPT} learns a transferable prompt by distilling from multiple task-specific prompts. XPrompt \citep{ma-etal-2022-xprompt} empirically demonstrates the negative impact of trained prompt tokens and proposes a hierarchical structured pruning for a trained soft prompt, which can be seen as a post-training method. 
DPT \citep{xiao-etal-2023-decomposed} adopts the product of two low-rank matrices to approximate the soft prompt. Residual Prompt Tuning \citep{ResidualPrompt} reparameterizes soft prompt embeddings with a shallow feed-forward network and residual connection, while ADePT uses the feed-forward network to learn input token embedding offsets—a distinct objective. DePT \citep{shidept} first discovers that the input token embedding offsets can enhance the performance of PT and uses short soft prompts to accelerate inference. ADePT produces adaptive input token embedding offsets via shallow token-shared networks, addressing the limitations of DePT mentioned in Section \ref{sec:intro}. With a comparable number of trainable parameters, ADePT can outperform both vanilla PT and DePT.

\section{Method}
In this section, we revisit the preliminaries of PT and DePT (Section \ref{sec:background_PT}) and analyze the limitations of DePT (Section \ref{sec:limitations}). Then, we introduce our proposed Adaptive Decomposed Prompt Tuning (ADePT) in Section \ref{sec:background_ADePT}, followed by a theoretical analysis of ADePT in Section \ref{theoretical analysis}.

\subsection{Preliminaries: Prompt Tuning (PT) and Decomposed Prompt Tuning (DePT)}
\label{sec:background_PT}
\textbf{Prompt Tuning(PT)}. Let $\mathcal{D} = \{\bm{X_i}, \bm{y_i}\}_{i=1}^{N}$ be the training dataset of the target downstream task $\mathcal{T}$, where $N$ is the number of the training data. Given a PLMs with parameters $\Theta$, each input $\bm{X}_i $ is first mapped to token embeddings $\bm{E}_i \in \mathbb{R}^{s \times d}$ by tokenizer and embedding layer, where $s$ denotes the maximum length of input tokens and $d$ denotes the dimension of the input token embeddings. The objective of PT is to learn a soft prompt $\bm{P} \in \mathbb{R}^{l \times d}$ to enable the adaptation of a PLM to the target downstream task. Here, $l$ denotes the length of the soft prompt. The soft prompt $\bm P$ is prepended to the input token embeddings $\bm E$. The loss of PT for the target downstream task is formulated as:
\begin{equation}
    \mathcal{L}_{\mathrm{PT}}=-\sum_i\log P(\bm{y}_i\mid\left[\bm{P},\bm{E}_i\right];\Theta),
\end{equation}
where $\mathcal{L}_{\mathrm{PT}}$ is the loss function only optimized with regard to the soft prompt $\bm{P}$.

\textbf{Decomposed Prompt Tuning (DePT).} DePT adapts PLMs to the new target downstream task via a short soft prompt $\bm{P}_{s_1} \in \mathbb{R}^{l_{s_1} \times d}$ and a pair of low-rank matrices, \textit{i.e.}, $\bm{A} \in R^{s \times r_s}$ and $\bm{B} \in \mathbb{R}^{r_s \times d}$, where $l_{s_1}$ denotes the length of short soft prompt and $r_s \ll \min(s,d)$ denotes the maximum rank of matrices $\bm{A}$ and $\bm{B}$. Similar to PT, the short soft prompt of DePT is prepended to the frozen input token embeddings. The product of low-rank matrices of DePT is used as the offsets of the input word embeddings.
The loss of DePT for the target downstream task is formulated as:
\begin{equation}
    \mathcal{L}_{\mathrm{DePT}}=-\sum_i\log P(\bm{y}_i\mid\left[\bm{P}_{s_1},\bm{E}_i + \bm{A} \bm{B} \right];\Theta),
\end{equation}
where the loss function $\mathcal{L}_{\mathrm{DePT}}$ is optimized only with respect to the short soft prompt $\bm{P}_{s_1}$ and the pair of low-rank matrices $\bm{A}$ and $\bm{B}$.
\begin{table}[!t]
\caption{The comparisons between DePT and ADePT on RTE and BoolQ tasks. All results are based on the T5-base model. The numbers within ``DePT()'' indicate the amount of cyclic left shift applied to the position-based input embedding offsets of DePT.}
\centering
\resizebox{0.9\textwidth}{!}{
\begin{tabular}{lccccca}
\toprule
 & DePT (0)& DePT (50) & DePT(100) & DePT(150) & DePT(200)  & ADePT \\
\midrule
RTE & 79.1 &  78.4 & 79.9 & 79.1 & 78.4  & 82.0\\
BoolQ & 78.4 & 74.7 & 73.1 & 70.9 & 73.9 & 80.2\\
\bottomrule
\end{tabular}
}
\label{table:position}
\end{table}

\begin{table}[!t]
\caption{The Mean and Variance of input token embeddings and embedding offsets of DePT and ADePT on RTE and BoolQ tasks. All results are based on the T5-base model. The Mean and Variance are calculated from the tokens of the entire training dataset. The ``mean()'' and ``variance()'' refer to the corresponding task inside the parentheses. }
\centering
\resizebox{\textwidth}{!}{
\begin{tabular}{lcccc}
\toprule
 & Mean (RTE) & Variance (RTE) & Mean (Boolq) & Variance (Boolq) \\
\midrule
Input Embeddings & 6.07 & 16.29 & 7.93 & 9.98 \\
The offset (absolute value) of DePT & 0.01 & 0.06 & 0.02 & 0.01 \\
The offset (absolute value) of ADePT & 8.31 & 5.45 & 6.09 &  3.70\\
\bottomrule
\end{tabular}
}
\label{table:mean_and_variance}
\end{table}

\subsection{The Limitations of DePT}
\label{sec:limitations}
In this section, we explore several key factors that limit the performance of DePT. 

Let $\bm{A} \bm{B} = \Delta \bm{E}$, the updated input token embeddings of DePT is formulated as $[\bm{e}_1 + \Delta \bm{e}_1, \bm{e}_2 + \Delta \bm{e}_2, \cdots, \bm{e}_s + \Delta \bm{e}_s]$. For a downstream task, the input token embeddings $\bm{E}$ vary. However, the pair of low-rank matrices are fixed after adaptation, meaning that the input token embedding offsets $\Delta \bm{E}$ are position-based. Assume there are input token embeddings $[\bm{a}, \bm{b}, \bm{c}]$ (here, we omit the right padding), the updated token embeddings are formulated as $[\bm{a} + \Delta \bm{e}_1, \bm{b} + \Delta \bm{e}_2, \bm{c} + \Delta \bm{e}_3]$. Assume we have a meaningless sequence that does not affect the prediction performance, denoted as $[\bm{t}_1, \bm{t}_2]$. Prepending the meaningless sequence to input token embeddings $[\bm{a}, \bm{b}, \bm{c}]$, the updated input token embeddings are $[\bm{t}_1 + \Delta \bm{e}_1, \bm{t}_2 + \Delta \bm{e}_2, \bm{a} + \Delta \bm{e}_3, \bm{b} + \Delta \bm{e}_4, \bm{c} + \Delta \bm{e}_5]$. We can observe that offsets of $[\bm{a}, \bm{b}, \bm{c}]$ are different. The offsets of the former is $[\Delta \bm{e}_1, \Delta \bm{e}_2, \Delta \bm{e}_3]$, while the offsets of the latter is $[\Delta \bm{e}_3, \Delta \bm{e}_4, \Delta \bm{e}_5]$. Just adding a meaningless sequence $[\bm{t}_1, \bm{t}_2]$ that does not affect the prediction results, the token embedding offsets for $[\bm{a}, \bm{b}, \bm{c}]$ become different. These position-dependent embedding offsets make the embeddings of each token non-unique in one task, which may be detrimental to the model performance. To simulate this scenario, we design an ideal experiment where we cyclically left shift the column vectors of $\Delta \bm{E}$ and then test the model performance, as shown in Table \ref{table:position}. Let $\Delta \bm{E}'$ be the cyclic left shift of $\Delta {\bm E}$ by $j$ positions, defined as $
\Delta \bm{E}' = [\Delta \bm{e}_{1+j}, \Delta \bm{e}_{2+j}, \cdots, \Delta \bm{e}_s, \bm{e}_1, \cdots, \bm{e}_j].$
For example, on RTE, the performance of DePT is decreased by 0.7 points after cyclically left-shifting the position-based embedding offsets by 50 positions, and increased by 0.8 points after shifting by 100 positions. On BoolQ, the performance of DePT is worse than the original after cyclically left-shifting the position-based embedding offsets. Table \ref{table:position} shows that position-based embedding offsets in DePT can cause unstable prediction performance across different positions.

Table \ref{table:mean_and_variance} reports the mean and variance of elements in input token embedding $\bm{e}$ and the elements in embedding offset $\Delta \bm{e}$ from DePT across two entire training datasets, \textit{i.e.}, RTE and BoolQ tasks. All results are calculated by their absolute values. We can observe that the mean and variance of elements in $\Delta \bm{e}$ are only a few percent of those of elements in $\bm{e}$. For example, on the RTE task, the mean absolute value of elements in $\Delta \bm{e}$ is only 0.01, while the mean absolute value of elements in $\bm{e}$ is 6.07. This implies that DePT makes only minor changes to the input token embedding space, which may result in its inability to map the input token embeddings to the appropriate embedding space. This is because, for token $\bm e$, its offset can be any position $\Delta \bm{e}_i$, where $1 \le i \le s$. The requirement that tokens within the same task should be unique causes the offsets of DePT to become extremely small, leading to the sub-optimization of DePT. The elements in the embedding offsets of ADePT have a much larger range of values than that of DePT. The optimal embedding space may lie outside the range of DePT, whereas ADePT may be able to access this embedding space.


\subsection{Our method: Adaptive Decomposed Prompt Tuning (ADePT)}
\label{sec:background_ADePT}
The limitations of DePT lie in its position-based token embedding offsets. To address this issue, the input token embedding offsets should be tailored for model input, and the corresponding embedding for each token should be unique after being offset. We find that making the input token embedding offsets $\Delta \bm{E}$ a function of the input token embedding $\bm{E}$ can meet such requirements. For an input token embedding $\bm{e}$, we want to get a function $f()$, which can produce the offset of this input token embedding, namely, $f(\bm{e})$. For the input token embeddings $\bm{E} = [\bm{e}_1, \bm{e}_2, \cdots, \bm{e}_s]$, the updated input token embedding can be formulated as $\bm {E}' = \bm{E} + \Delta \bm{E} = [\bm{e}_1 + f(\bm{e}_1), \bm{e}_2 + f(\bm{e}_s), \cdots, \bm{e}_s + f(\bm{e}_s)]$. For example, the input token embeddings $[\bm{a}, \bm{b}, \bm{c}]$ can be updated as $[\bm{a} + f(\bm{a}), \bm{b} + f(\bm{b}), \bm{c} + f(\bm{c})]$. Prepending the meaningless sequence $[\bm{t}_1, \bm{t}_2]$ to input token embeddings $[\bm{a}, \bm{b}, \bm{c}]$, the updated input token embeddings are $[\bm{t}_1 + f(\bm{t}_1), \bm{t}_2 + f(\bm{t}_2), \bm{a} + f(\bm{a}), \bm{b} + f(\bm{b}), \bm{c} + f(\bm{c})]$. We can observe that the offsets for $[\bm{a}, \bm{b}, \bm{c}]$ are the same in this two scenario, \textit{i.e.}, $[f(\bm{a}), f(\bm{b}), f(\bm{c})]$. Therefore, if such a function $f$ exists, we can achieve input token embedding offsets that are tailored for model input, and the embedding for each token is unique within a task. 

To avoid increasing the number of trainable parameters, we use a shallow and narrow feed-forward neural network to approximate the function $f$. Thus, we propose Adaptive Decomposed Prompt Tuning (ADePT), which can offset the token embeddings adaptively based on the model input. We implement the shallow token-shared feed-forward neural network by a two-layer multi-layer perceptron (MLP). It consists of a down-projection matrix $\bm{W}_{\text{down}} \in \mathbb{R}^{d \times r}$ and a up-projection matrix $\bm{W}_{\text{up}} \in \mathbb{R}^{r \times d}$, and a down-projection bias $\bm{b}_1 \in \mathbb{R} ^ {r} $ and a up-projection bias $\bm{b}_2 \in \mathbb{R} ^ {d} $. Here, $r$ is the bottleneck size of the MLP. The updated input token embeddings by the shallow token-shared feed-forward neural network are formulated as:
\begin{equation}
    \bm{E}'_i = \bm{E}_i + \text{ReLU} (\bm{E}_i \bm{W}_{\text{down}} + \bm{b}_1) \bm{W}_{\text{up}} + \bm{b}_2.
\end{equation}
To ensure that the number of trainable parameters does not exceed that of the vanilla PT, we use a short soft prompt $\bm{P}_{s_2} \in \mathbb{R}^{l_{s_2} \times d}$, similar to DePT. The loss of ADePT is formulated as:
\begin{equation}
    \mathcal{L}_{\mathrm{ADePT}}=-\sum_i\log P(\bm{y}_i\mid\left[\bm{P}_{s_2},\bm{E}'_i \right];\Theta),
\end{equation}
where the loss function $\mathcal{L}_{\mathrm{ADePT}}$ is optimized only with respect to the short soft prompt $\bm{P}_{s_2}$ and the parameters of the feed-forward neural network $\bm{W}_{\text{down}}$, $\bm{W}_{\text{up}}$, $\bm{b}_1$ and $\bm{b}_2$.


\subsection{Theoretical analysis}
\label{theoretical analysis}
In this section, inspired by \citet{promptheory}, we provide a theoretical analysis towards ADePT.

The multi-head self-attention layer serves as a crucial component in each transformer layer. We analyze how PT and ADePT affect the first transformer layer. To simplify the analysis, let us consider a single head self-attention $\mathcal{H}$ in the first layer, which is parameterized by $\bm{W}_Q^{\mathcal{H}}, \bm{W}_K^{\mathcal{H}}, \bm{W}_Q^{\mathcal{H}} \in \mathbb{R}^{d \times d_\mathcal{H}}$. Given a input sequence embeddings $\bm{E} = (\bm{e}_1, \bm{e}_2, \dots, \bm{e}_s) \in \mathbb{R}^{s \times d}$ with each $\bm{e} \in \mathbb{R}^{d}$, the output of a query vector $\bm{e}_i$ passing through the single-head self-attention $\mathcal{H}$ in the first layer is formulated as:
\begin{equation}
    \bm{o}_i = \text{Attention}\left(\bm{e}_i \bm{W}_Q^{\mathcal{H}}, \bm{E} \bm{W}_K^{\mathcal{H}} , \bm{E} \bm{W}_V^{\mathcal{H}}  \right) = \text{Softmax}\left( \left(\bm{e}_i \bm{W}_Q^{\mathcal{H}}\right)\left(\bm{E}\bm{W}_K^{\mathcal{H}} \right)^T\right) \bm{E} \bm{W}_V^{\mathcal{H}},
\end{equation}
where the scaling constant $\sqrt{d_\mathcal{H}}$ is ignored for notation convenience.

For the vanilla PT with the soft prompt $\bm{P} = [\bm{p}_1, \bm{p}_2, \dots, \bm{p}_l] \in \mathbb{R}^{l \times d}$, the output of a query vector $\bm{e}_i$ passing through the single-head self-attention $\mathcal{H}$ in the first layer is formulated as:
\begin{align}
\begin{aligned}
    & \bm{o}_i^{\text{PT}}  =  \text{Attention}\left( \bm{e}_i \bm{W}_Q^\mathcal{H}, \text{concat}[\bm{P},\bm{E}] \bm{W}_K^\mathcal{H}, \text{concat}[\bm{P},\bm{E}] \bm{W}_V^\mathcal{H} \right)\\
    & = \underbrace{\sum_{k=1}^{l} \bm{A}_{ik} \bm{p}_k \bm{W}_V^{\mathcal{H}}}_{bias} + \underbrace{\left(1 - \sum_{k=1}^{l} \bm{A}_{ik}\right)}_{scale}\bm{o}_i, \\
    \text{with}\quad \bm{A}_{ik} &= \frac{\exp\left(\bm{e}_i \bm{W}_Q^{\mathcal{H}} \left(\bm{p}_k \bm{W}_K^{\mathcal{H}}\right)^T \right)}{\sum_{k=1}^{l}\exp\left(\bm{e}_i \bm{W}_Q^\mathcal{H}  \left(\bm{p}^k \bm{W}_K^\mathcal{H}\right)^T \right) + \sum_{j=1}^{s} \exp \left(\bm{e}_i \bm{W}_Q^{\mathcal{H}} \left(\bm{e}_j \bm{W}_K^\mathcal{H}\right)^T \right)},
    \label{eq:PTtheory}
\end{aligned}
\end{align}
where $\bm{A}_{ik}$ is the attention score assigned to the prefix vector $\bm{p}_k$ for $\bm{e}_i$. Thus, in the first transformer layer, PT cannot affect the relative attention patterns across the content and it only scales the attention patterns down while adding a constant bias to the original output $\bm{o}_i$ \citep{promptheory}.


For our proposed ADePT with the soft prompt $\bm{P} = [\bm{p}_1, \bm{p}_2, \dots, \bm{p}_l] \in \mathbb{R}^{l \times d}$ and feed-forward neural network $f$, the output of a query vector $\bm{e}_i$ passing through the single-head self-attention $\mathcal{H}$ in the first layer is formulated as:
\begin{align}
\begin{aligned}
   & \bm{o}_i^{\text{ADePT}}  =  \text{Attention}\left(\left( \bm{e}_i + f\left(\bm{e}_i\right) \right)\bm{W}_Q^\mathcal{H} , \text{concat}[\bm{P},\bm{E} + f\left(\bm{E}\right)  ] \bm{W}_K^\mathcal{H} , \text{concat}[\bm{P},\bm{E} + f\left(\bm{E}\right) ] \bm{W}_V^\mathcal{H}  \right)\\
    & \quad \quad =  \sum_{k=1}^{l} \bm{A}_{ik} \bm{p}_k \bm{W}_V^{\mathcal{H}}  \\ 
     & \quad \quad + (1 - \sum_{k=1}^{l}  \bm{A}_{ik}) \text{Softmax}\left( \left(\left( \bm{e}_i + f\left(\bm{e}_i\right) \right) \bm{W}_Q^{\mathcal{H}} \right)\left(\left(\bm{E}+f\left(\bm{E}\right) \right) \bm{W}_K^{\mathcal{H}}  \right)^T\right) \left(\bm{E} + f\left(\bm{E}\right) \right)\bm{W}_V^{\mathcal{H}} , \\
    &  \text{with}\quad \bm{A}_{ik} = \frac{\exp\left( \left(\left(\bm{e}_i + f\left(\bm{e}_i\right) \right) \bm{W}_Q^{\mathcal{H}}  \right) \left(\bm{p}_k \bm{W}_K^{\mathcal{H}}\right)^\top \right)}{B},\\
    & \quad \quad B = \sum_{k=1}^{l} \exp\left(\left(\left(\bm{e}_i + f\left(\bm{e}_i\right) \right) \bm{W}_Q^\mathcal{H} \right)\left(\bm{p}_k \bm{W}_K^\mathcal{H}\right)^T \right) \\
    & \quad \quad \quad + \sum_{j=1}^{s} \exp\left(\left(\left(\bm{e}_i + f\left(\bm{e}_i\right) \right) \bm{W}_Q^{\mathcal{H}} \right) \left(\left(\bm{e}_j+f(\bm{e}_j)\right) \bm{W}_K^\mathcal{H}\right)^T \right).
    \label{eq:ADePTtheory}
\end{aligned}
\end{align}
Hence, in the first transformer layer, ADePT can change the original relative attention patterns and add a bias dependent on the input, which makes ADePT have more expressive power than PT.




\section{Experiments and Results}
\subsection{EXPERIMENTAL SETUP}
\textbf{Tasks and Models.} We conduct extensive experiments to validate our proposed ADePT. We consider four benchmarks and 4 other datasets: (1) GLUE \citep{wang-etal-2018-glue} benchmark, which includes MNLI \citep{williams-etal-2018-broad}, QQP\footnote{\url{https://www.quora.com/q/quoradata/}}, QNLI \citep{rajpurkar-etal-2016-squad}, SST-2 \citep{socher-etal-2013-recursive}, STS-B \citep{cer-etal-2017-semeval}, MRPC \citep{dolan-brockett-2005-automatically}, RTE \citep{giampiccolo-etal-2007-third} and CoLA \citep{warstadt2019neural_cola};
(2) SuperGLUE benchmark \citep{wang2019superglue}, which includes MultiRC \citep{khashabi-etal-2018-looking}, BoolQ \citep{clark-etal-2019-boolq}, WiC \citep{pilehvar-camacho-collados-2019-wic}, WSC \citep{levesque2012winograd_wnli}, CB \citep{de2019commitmentbank} and ReCoRD \citep{zhang2018record}; 
(3) MRQA 2019 Shared Task \citep{fisch-etal-2019-mrqa}, which includes Natural Questions \citep{kwiatkowski-etal-2019-natural}, HotpotQA \citep{yang-etal-2018-hotpotqa}, SearchQA \citep{dunn2017searchqa} and NewsQA \citep{trischler-etal-2017-newsqa};
(4) MBPP benchmark \citep{MBPP}, which is a code generation task;
(5) other datasets, which includes WinoGrande \citep{10.1145/3474381}, Yelp-2 \citep{10.5555/2969239.2969312}, SciTail \citep{Khot_Sabharwal_Clark_2018} and PAWS-Wiki \citep{zhang-etal-2019-paws}. Following~\citep{asai-etal-2022-attempt,MPT,shidept}, we evaluate our proposed ADePT on all datasets for the T5-base model (220M) \citep{t5}, except for MBPP and ReCoRD. For the T5-3B model \citep{t5}, we focus on large and challenging datasets (\textit{i.e.}, MNLI, ReCoRD, Natural Questions, HotpotQA, SearchQA, and NewsQA) to differentiate the performance of various PEFT methods. 
For the decoder-only PLMs (\textit{i.e.}, CodeGen-350M \citep{nijkamp2023codegen} and Llama3-8B \citep{llama3}), we evaluate our proposed method ADePT on MBPP benchmark.


\textbf{Baselines.} To evaluate our proposed ADePT, we compare it with five types of fine-tuning methods: (1) full fine-tuning (FT), which optimizes all the model parameters; (2) the vanilla PT \citep{lester-etal-2021-power}, where target prompt vectors are initialized with randomly sampled top vocabularies; (3) the variants of PT using additional transfer or multi-task learning, including SPoT \citep{vu-etal-2022-spot}, ATTEMPT \citep{asai-etal-2022-attempt}, and MPT \citep{MPT}; (4) the variants of PT using input token embedding offsets, \textit{i.e.}, DePT \citep{shidept}; (5) state-of-the-art PEFT methods including  Adapter\citep{houlsby2019parameter},  AdapterDrop \citep{ruckle-etal-2021-adapterdrop}, BitFit \citep{zaken2021bitfit},  HyperFomer \citep{mahabadi2021parameter}, HyperDecoder \citep{ivison-peters-2022-hyperdecoders}, P-tuning \citep{liu2021gpt}, LoRA \citep{hu2021lora},  LST \citep{sunglst}, and their multi-task learning variants.

\textbf{Implementation Details.} 
Following \citet{shidept}, we use $100$ learnable virtual tokens as the soft prompt of PT. For our proposed ADePT, we adjust the hyperparameters to maintain an equivalent number of trainable parameters as PT. For instance, in the T5-base model, the token embedding dimension $d$ is $768$, so the number of trainable parameters is $l \times d = 100 \times 768 = 76800$. Following \citet{shidept}, we search the length of soft prompt from $20$, $40$, $60$, and $80$. For ADePT, if using $60$ virtual tokens for soft prompt, the $d_r$ is got by solving the unequal equation $60 \times 768 + 2 \times d_r \times 768 + d_r + 768 \leq 76800$. Thus, the $d_r \leq 19.49$ and $d$ is set to $19$ because the $d$ is the integer. According to this calculation method, the corresponding $d_r$ values for soft prompt lengths of $20$, $40$, $60$ and $80$ are $39$, $29$, $19$, and $9$, respectively. For a fair comparison of the T5-base model, we directly quote performance metrics from published papers \citep{karimi2021compacter,mahabadi2021parameter,asai-etal-2022-attempt,MPT,sunglst,shidept}. For T5-3B model, we consistently use $60$ virtual tokens and bottleneck size $r=19$. Due to the lack of experimental results, for a fair comparison with the T5-3B model, we reproduce the experiments of the vanilla PT and DePT. For decoder-only PLMs, following \citet{promptback}, we use $10$ virtual tokens for PT, $7$ virtual tokens and rank $r_s = 3$ for DePT, $7$ virtual tokens and bottleneck size $r=1$ for ADePT, and rank $16$ for LoRA. For small datasets ($<70,000$ training samples) based on T5 model, we follow the learning strategy of \citet{shidept}: we search the learning rate for the soft prompt from $3e-1$, $4e-1$, $5e-1$, and for the feed-forward neural network from $1e-4$, $1e-5$. For large datasets ($>70,000$ training samples) based on T5 model, we use learning rate 3e-1 for the soft prompt and $1e-4$ for the feed-forward neural networks. For the MBPP benchmark, following \citet{promptback}, we use learning rates of $1e-3$ for the prompting-style tuning method, $1e-4$ for LoRA. 


\begin{table}[!t]
\centering
\caption{
The experimental results on GLUE and SuperGLUE benchmarks, with the associated size of trainable parameters. All results are based on the T5-base model. We report Pearson correlation for STS-B, F1 for MultiRC (Multi), and accuracy for other tasks as test metrics.
}

\vspace{0.1em}
\label{table: glue_and_super_glue}
\resizebox{1.0\textwidth}{!}{
\begin{threeparttable}
\addtolength{\tabcolsep}{-4.25pt}  
\begin{tabular}{lrccccccccaccccca}
\toprule
\multirow{2}{*}{\bf Method} & \multirow{2}{*}{\bf \#Para}    &\multicolumn{9}{c}{\textbf{GLUE}}           & \multicolumn{6}{c}{\textbf{SuperGLUE}} \\ 
                                                     \cmidrule(lr){3-11}                           \cmidrule(lr){12-17}
                 &       & MNLI & QQP  & QNLI & SST-2  & STS-B & MRPC  & RTE   & CoLA  & \cg Mean &  Multi     & Bool  & WiC   & WSC  & CB  & \cg Mean \\  \midrule
\multicolumn{17}{c}{\bf \textit{Single-Task Learning}} \\ \midrule
Full Finetuning$^1$\quad  & 220M  & 86.8 & 91.6 & 93.0 & 94.6   & 89.7  & 90.2  & 71.9  & 61.8  & \cg 84.9 &  72.8      & 81.1  & 70.2  & 59.6 & 85.7 & \cg 73.9 \\ 
Adapter$^1$      & 1.9M  & 86.5 & 90.2 & 93.2 & 93.8   & 90.7  & 85.3  & 71.9  & 64.0  & \cg 84.5 &  75.9      & 82.5  & 67.1  & 67.3 & 85.7 & \cg 75.7 \\
AdapterDrop$^1$  & 1.1M  & 86.3 & 90.2 & 93.2 & 93.6   & 91.4  & 86.3  & 71.2  & 62.7  & \cg 84.4 &  72.9      & 82.3  & 68.3  & 67.3 & 85.7 & \cg 75.3 \\
BitFit$^1$       & 280K  & 85.3 & 90.1 & 93.0 & 94.2   & 90.9  & 86.8  & 67.6  & 58.2  & \cg 83.3 &  74.5      & 79.6  & 70.0  & 59.6 & 78.6 & \cg 72.5  \\
LoRA$^2$         & 3.8M  & 86.3 & 89.0 & 93.2 & 94.3   & 90.9  & 90.1  & 75.5  & 63.3  & \cg 85.3 &  72.6     &  81.3 &  68.3 & 67.3   & 92.9  & \cg 76.5   \\
LST$^2$          & 3.8M  & 85.6 & 88.8 & 93.3 & 94.1   & 90.7  & 90.4  & 71.9  & 58.1  & \cg 84.1 &  --        & --    & --    & --   & --   & \cg -- \\
PT$^4$           & 76.8K & 83.4 & 90.2 & 93.1 & 91.9   & 90.2  & 90.1  & 78.8  & 60.7  & \cg 84.8 &  65.7      & 63.7  & 50.8  & 51.9 & 67.9 & \cg 60.0  \\
%
DePT$^4$     & 76.8K & 85.0 & 90.4 & 93.2 & 94.2   & 90.8  & 90.7  & 79.1  & 63.8  & \cg 85.9 &  74.3      & 79.3  & 68.7 & 67.3  & 92.9 & \cg 76.5 \\
ADePT (ours)    & 76.1K & 85.7 & 90.4 & 93.2 & 94.0   & 90.9  & 91.2  & 82.0  & 65.5  & \cg \textbf{86.6} &  74.6      & 80.2  & 68.7 & 67.3  & 96.4 & \cg \textbf{77.4} \\
\midrule
\multicolumn{17}{c}{\bf \textit{Additional Transfer Learning or Multi-Task Learning}} \\
\midrule
Full Fine-tuning (m)$^1$  & 28M      & 85.7 & 91.1 & 92.0 & 92.5 & 88.8 & 90.2 & 75.4 & 54.9 & \cg 83.8 &  74.4     & 81.1      & 70.0  & 71.2   &  85.7 & \cg 76.1  \\
Adapter (m) $^1$      & 1.8M     & 86.3 & 90.5 & 93.2 & 93.0 & 89.9 & 90.2 & 70.3 & 61.5 & \cg 84.4 &  72.6    & 82.3      &  66.5 & 67.3   &  89.3  &  \cg 75.6  \\
HyperFormer (m) $^1$  & 638K     & 85.7 & 90.0 & 93.0 & 94.0 & 89.7 & 87.2 & 75.4 & 63.7 & \cg 84.8 & 72.9     & 82.5      &  69.0 & 67.3   & 85.7  & \cg 75.4 \\
HyperDecoder (m) $^1$ & 1.8M     & 86.0 & 90.5 & 93.4 & 94.0 & 90.5 & 87.7 & 71.7 & 55.9 & \cg 83.7 & 70.4     &  78.8     & 67.1  & 61.5   & 82.1  & \cg 72.0 \\
SPoT (t) $^1$         & 76.8K  & 85.4 & 90.1 & 93.0 & 93.4 & 90.0 & 79.7  & 69.8  & 57.1 & \cg 82.3    &  74.0      & 77.2  & 67.0  & 50.0 & 46.4 & \cg 62.9 \\
ATTEMPT (t) $^1$      & 232K   & 84.3 & 90.3 & 93.0 & 93.2 & 89.7 & 85.7  & 73.4  & 57.4 & \cg 83.4    &  74.4      & 78.8  & 66.8 & 53.8  & 78.6 & \cg 70.5 \\ 
MPT (t) $^3$          & 77.6K  & 85.9 & 90.3 & 93.1 & 93.8 & 90.4 & 89.1  & 79.4  & 62.4 & \cg 85.6    &  74.8      & 79.6  & 69.0 & 67.3  & 79.8 & \cg 74.1 \\
ATTEMPT (m)$^3$     & 96K  & 83.8 & 90.0 & 93.1 & 93.7 & 90.8 & 86.1  & 79.9 & 64.3 &  \cg 85.2  & 74.4    & 78.5  & 66.5 &   69.2 & 82.1 & \cg 74.1 \\ 
MPT (m)$^3$         & 10.5K & 84.3 & 90.0 & 93.0 & 93.3 & 90.4 & 89.2  & 82.7 & 63.5 &  \cg 85.8  & 74.8    & 79.2  & 70.2 &   67.3 & 89.3 & \cg 76.1 \\
\bottomrule
\end{tabular}
$^1$ sourced from \citep{asai-etal-2022-attempt}.
$^2$ sourced from \citep{sunglst}. 
$^3$ sourced from \citep{MPT}. 
$^4$ sourced from \citep{shidept}.
(m) represents additional multi-task training.
(t) represents additional transfer learning.
\end{threeparttable}
}
\vspace{-1em}
\end{table}

\begin{table}[!t]
\centering
\caption{The experimental results on MRQA 2019 Shared Task and other datasets with the associated size of trainable parameters. All results are based on the T5-base model. We report the F1 for MRQA tasks and accuracy for other datasets as test metrics. The results are averaged over three runs and the subscripts denote standard deviation. All baseline results are quoted from \citep{shidept}.}
\label{table:mrqa}
\vspace{0.5em}
\resizebox{\textwidth}{!}{
\begin{tabular}{lcccccacccca}
\toprule 
\multirow{2}{*}{\bf Method} & \multirow{2}{*}{\bf \#Para} & \multicolumn{5}{c}{\bf MRQA}         & \multicolumn{5}{c}{\bf Others}  \\ 
\cmidrule(lr){3-7} \cmidrule(lr){8-12} 
& & NQ & HP & SQA & News & Mean & WG & Yelp & SciTail & PAWS & Mean \\ 
\midrule
Full Fine Tuning            & 220M   & 75.1 & 77.5 & 81.1 & 65.2 & 74.7 & 61.9 & 96.7 & 95.8 & 94.1 & 87.1 \\ 
Adapter             & 1.9M   & 74.2 & 77.6 & 81.4 & 65.6 & 74.7 & 59.2 & 96.9 & 94.5 & 94.3 & 86.2 \\
BitFit                 & 280K   & 70.7 & 75.5 & 77.7 & 64.1 & 72.0 & 57.2 & 94.7 & 94.7 & 92.0 & 84.7 \\
LoRA                   & 3.8M   & 72.4 & 62.3 & 72.5 & 56.9 & 66.0 & 58.2 & 97.1 & 94.7 & 94.0 & 86.0 \\
PT                     & 76.8K  & 67.9 & 72.9 & 75.7 & 61.1 & 69.4 & 49.6 & 95.1 & 87.9 & 55.8 & 72.1 \\
SPoT                   & 76.8K  & 68.2 & 74.8 & 75.3 & 58.2 & 69.1 & 50.4 & 95.4 & 91.2 & 91.1 & 82.0 \\
ATTEMPT                & 232K   & 70.4 & 75.2 & 77.3 & 62.8 & 71.4 & 57.6 & 96.7 & 93.1 & 92.1 & 84.9 \\
MPT                    & 77.6K  & $72.0_{0.1}$ & $75.8_{0.1}$ & $77.2_{0.1}$ & $63.7_{0.1}$ & $72.2$ & $56.5_{0.9}$ & $96.4_{0.0}$ & $95.5_{0.1}$ & $93.5_{0.1}$ & $85.5$ \\
DePT                   & 76.8K  & $73.2_{0.1}$ & $76.8_{0.3}$ & $77.6_{0.2}$ & $64.4_{0.1}$ & $73.0$ & $59.0_{0.2}$ & $96.8_{0.1}$ & $95.6_{0.2}$ & ${93.7}_{0.1}$ & $86.3$ \\
ADePT (ours)                 & 76.1K  & ${73.9}_{0.0}$ & ${77.1}_{0.1}$ & ${78.7}_{0.1}$ & ${64.7}_{0.1}$ & ${73.6}$ & ${59.1}_{0.9}$ & ${96.8}_{0.0}$ & ${95.9}_{0.3}$ & $93.7_{0.2}$ & ${86.4}$ \\
\bottomrule
\end{tabular}
}
\vspace{-0.5em}
\end{table}

\subsection{Results based on T5-base model}

\textbf{\#1 Performance on GLUE and SuperGLUE benchmarks.}

As demonstrated in Table \ref{table: glue_and_super_glue}, our proposed ADePT surpasses leading PEFT methods, including Adapter, LoRA, BitFit, and LST, on the GLUE and SuperGLUE benchmarks, while utilizing the least trainable parameters. ADePT outperforms the vanilla PT while using comparable trainable parameters and less inference time. ADePT also outperforms the variants of PT using additional transfering learning, including SPoT, ATTEMPT and MPT while not requiring the complicated training and storage of soft prompts for source tasks. Remarkably, ADePT outperforms DePT, demonstrating that the adaptive input token embedding offsets by token-shared feed-forward neural networks are better than the position-based input token embedding offsets. Moreover, ADePT can even outperform the full finetuning method and the PEFT methods using additional multi-task learning.

\textbf{\#2 Performance on MRQA 2019 Shared Task and other four datasets.}

Table \ref{table:mrqa} presents the performance of different PEFT methods in the MRQA dataset and four other tasks.
Despite having fewer parameters (76.1K) and faster inference (shorter soft prompt), ADePT shows a significant improvement of $6.1\%$ on MRQA and $19.8\%$ on the four other datasets over the vanilla PT. Also, ADePT surpasses the variants of PT using additional transfer learning, including SPoT, ATTEMPT and MPT on MRQA and the other four tasks. Furthermore, ADePT can consistently outperform DePT on MRQA and achieve comparable performance compared to DePT on the other four tasks. Compared to Adapter, ADePT can achieve comparable performance when only using $4.0\%$ trainable parameters. 

\begin{table}[!t]
\centering
\caption{
The experimental results on six large and challenging tasks with the associated size of trainable parameters. All results are based on the T5-3B model. We use F1 for Natural Questions, HotpotQA, SearchQA, NewsQA, and ReCoRD, and accuracy for MNLI as test metrics.
}
\label{table:T5_3B_F1}
\vspace{0.1em}
\resizebox{0.8\textwidth}{!}{
\begin{tabular}{lccccccca}
\toprule 
\bf Method & \bf \#Para   & NQ   & HP   & SQA  & News & MNLI & ReCoRD & Mean\\ 
\midrule
LoRA & 25.8M & \textbf{80.6} & \textbf{82.6} & \textbf{87.1} & \textbf{69.5} & \textbf{91.3} & 72.8 & \textbf{80.7}\\
PT  & 102.4K  & 77.5 & 80.8 & 84.5 & 67.7 & 90.7 & \underline{72.9} & 79.0\\
DePT & 101.4K  & 77.2 & 80.7 & 83.8 & 66.4 & 89.7 & 72.8 &  78.4\\
ADePT (ours) & 101.4K  & \underline{77.7} & \underline{80.9} & \underline{84.7} & \underline{67.8} & \underline{90.9} & \textbf{73.0} &  \underline{79.2}\\
\bottomrule
\end{tabular}
}
\vspace{-.5em}
\end{table}

\begin{table}[!t]
\centering
\caption{
The experimental results on six large and challenging tasks with the associated size of trainable parameters. All results are based on the T5-3B model. We use EM (Exact Match) for Natural Questions, HotpotQA, SearchQA, NewsQA, and ReCoRD as test metrics.
}
\label{table:T5_3B_EM}
\vspace{0.1em}
\resizebox{0.7\textwidth}{!}{
\begin{tabular}{lcccccca}
\toprule 
\bf Method & \bf \#Para   & NQ   & HP   & SQA  & News & ReCoRD & Mean\\ 
\midrule
LoRA & 25.8M & \textbf{69.7}& \textbf{67.6} & \textbf{82.5}& \textbf{55.1} &\underline{59.2} & \textbf{66.8}\\
PT  & 102.4K  & 65.4 & 65.4 & 79.2 & \underline{51.6} & \underline{59.2} & 64.2\\
DePT & 101.4K  & 65.0 & 65.4 & 78.3 & 48.9 & 59.1 &  63.3 \\
ADePT (ours) & 101.4K  & \underline{65.8} & \underline{65.5} & \underline{79.4} & 51.5 & \textbf{59.3}&  \underline{64.3}\\
\bottomrule
\end{tabular}
}
\vspace{-0.5em}
\end{table}

\begin{table}[!t]
  \centering
  \renewcommand{\arraystretch}{1.2} 
  \captionsetup{skip=5pt} 
   \caption{Performance comparison on MBPP benchmark. We report average $\textit{pass}@1$ scores on CodeGen-350M and Llama3-8B models.} 

   \resizebox{0.6\columnwidth}{!}{
  \begin{tabular}{cccc}
    \toprule
   \multirow{2}{*}{\bf Model} & \multirow{2}{*}{\bf Method} & \multirow{2}{*}{\bf \#Para}  & \bf Code Generation\\
    & & & \bf MBPP \\
   \midrule
     
     \multirow{6}{*}{CodeGen-350M}
    & LoRA & 1.3M & \textbf{20.32}\\ 
     & PT & 10.2K &  16.12\\ 
     & DePT & 10.4K &  16.83\\
     & ADePT(ours) & 10.2K &  \underline{17.86}\\
    
    \hline
  

    
     
\hline
     \rule{0pt}{3ex}
     
     \multirow{6}{*}{Llama3-8B}
     & LoRA & 9.4M  & \textbf{49.08}\\ 
     & PT & 41.0K  & 18.27\\ 
     & DePT & 42.7K &  42.50\\
      & ADePT (ours) & 41.0k  & \underline{43.22}\\     \bottomrule
  \end{tabular}
}
\vspace{-1em}
  \label{tab:code}
\end{table}

\subsection{Results based on T5-3B model}
In this section, we evaluate our proposed ADePT the T5-3B model on six large and challenging tasks, including MNLI from the GLUE benchmark, ReCoRD from the SuperGLUE benchmark, and the MRQA 2019 Shared Task. 
Tables \ref{table:T5_3B_F1} and \ref{table:T5_3B_EM} present the experimental results of PT, DePT, and ADePT on six large and challenging tasks. DePT underperforms the vanilla PT across all tasks. There may be two reasons for this: (1) the position-based embedding offsets of DePT are harmful to the T5-3B model; (2) DePT is sensitive to hyperparameters, and the hyperparameters selected based on GLUE and SuperGLUE benchmarks are not conducive to the optimization of DePT. Both of these reasons indicate that the token embedding offsets based on position and the sharing of token embedding offsets among multiple tokens cause sub-optimization of PLMs, especially in billion-scale PLMs.  We can observe that ADePT almost achieves the optimal results across all tasks, indicating that the use of the feed-forward neural networks to learn adaptive embedding offset tailored for each token can still map the input embedding into better embedding space on billion-scale PLMs. Although our method does not perform as well as the LoRA on the T5-3B model, our method is the best among PT-style methods, and compared to LoRA, it can use significantly fewer parameters while flexibly switching parameters to adapt to different downstream tasks.

\subsection{Results based on Decoder-only PLMs}
We evaluate our proposed ADePT on decoder-only PLMs (i.e., CodeGen-350M model \citep{nijkamp2023codegen} and Llama3-8B model \citep{llama3}) through instruction tuning \citep{ouyang2022training}. We use MBPP benchmark, which is a Python program generation task \citep{MBPP}. Following \citet{promptback}, we use a 50-50 split for training and test. We report average $\textit{pass}@1$ scores to evaluate the performance, as shown in Table \ref{tab:code}. We can observe that ADePT performs best among PT-style methods, demonstrating its effectiveness. In the Llama3-8B model, the vanilla PT performs much worse than DePT and ADePT. This shows that the inability of PT to change the relative attention patterns limits its adaptation ability, whereas DePT and ADePT perform better because they can change the relative attention patterns. Also, in CodeGen-350M and Llama3-8B, the use of adaptive token embedding offsets helps ADePT perform better than DePT. Although ADePT performs slightly worse than LoRA on the MBPP benchmark, ADePT requires far fewer parameters than LoRA. More importantly, LoRA needs to merge weights, which typically limits it to only a single downstream task. In contrast, ADePT can flexibly switch between tasks and adapt to multiple downstream tasks simultaneously, which is a unique advantage of PT-style methods.

\subsection{Further Analysis}
\begin{table}[!t]
\centering
\caption{
Test results of longer soft prompt lengths using the T5-base model on the GLUE benchmark.
}
\vspace{0.1em}
\resizebox{\columnwidth}{!}{
\begin{tabular}{lccc}
\toprule
\bf Method  & \bf\#Para              & \bf Average Glue Performance   & \bf Inference samples per second (SST2) \\   
\midrule
PT (m=200)         &    153.6K    &  85.2 &	57.4 \\
DePT (m=120, r=60)  &   153.6K   &  86.0 &	 77.2  \\  
ADePT (m=120, r=39) (ours) &  152.9K     &  86.5 & 72.7  \\  
\bottomrule
\end{tabular}
}
\label{table:different_prompt_length2}
\vspace{-1em}
\end{table}
\textbf{Performance using longer soft prompt.} 
Table \ref{table:different_prompt_length2} compares the performance of vanilla PT, DePT, and ADePT under a comparable number of trainable parameters (corresponding to vanilla PT with a length of 200). The results show that ADePT outperforms both vanilla PT and DePT, demonstrating the effectiveness of its adaptive embedding offsets. In terms of inference speed, ADePT achieves speeds that exceed those of PT and match those of DePT. Although the additional shallow token-shared feed-forward neural network introduces some latency, the impact is minimal. It is worth noting that ADePT requires computing embedding offsets for each token in real-time for every model input, whereas DePT relies on fixed token embedding offsets. This real-time computation may introduce additional latency; however, if the embedding offsets obtained from the ADePT method are precomputed and added to the corresponding tokens in advance, this latency can be eliminated. Furthermore, since the addition operation $\mathbf{E} + \mathbf{A} \mathbf{B}$ in DePT cannot be avoided, the theoretical upper-speed limit of ADePT is expected to be faster than that of DePT, highlighting its potential to achieve both superior performance and faster inference speeds.



\section{Conclusion}
We propose a new parameter-efficient fine-tuning method, \textit{i.e.}, Adaptive Decomposed Prompt Tuning (ADePT), which consists of a short soft prompt and a shallow token-shared feed-forward neural network. The feed-forward neural network can learn a unique offset for each input token and map the input token embeddings into a better embedding space in a position-independent manner. Extensive experiments demonstrate that ADePT outperforms leading PEFT methods, including full fine-tuning, in certain scenarios. Further analysis demonstrates that ADePT exhibits a faster inference speed compared to the vanilla PT with a comparable number of parameters, while simultaneously achieving superior performance. We provide a theoretical analysis demonstrating that ADePT exhibits greater expressive power compared to PT and DePT. 

\section*{Acknowledgments}
This research was supported by National Natural Science Foundation of China (No.62476277), National Key Research and Development Program of China (NO.2024YFE0203200), CCF-ALIMAMA TECH Kangaroo Fund (No.CCF-ALIMAMA OF 2024008), and Huawei-Renmin University joint program on Information Retrieval. We also acknowledge the support provided by the fund for building worldclass universities (disciplines) of Renmin University of China and by the funds from Beijing Key Laboratory of Big Data Management and Analysis Methods, Gaoling School of Artificial Intelligence, Renmin University of China, from Engineering Research Center of Next-Generation Intelligent Search and Recommendation, Ministry of Education, from Intelligent Social Governance Interdisciplinary Platform, Major Innovation \& Planning Interdisciplinary Platform for the ``DoubleFirst Class'' Initiative, Renmin University of China, from Public Policy and Decision-making Research Lab of Renmin University of China, and from Public Computing Cloud, Renmin University of China.

\bibliographystyle{iclr2025_conference}
\bibliography{iclr2025_conference}

@article{MBPP,
  title={Program synthesis with large language models},
  author={Austin, Jacob and Odena, Augustus and Nye, Maxwell and Bosma, Maarten and Michalewski, Henryk and Dohan, David and Jiang, Ellen and Cai, Carrie and Terry, Michael and Le, Quoc and others},
  journal={arXiv preprint arXiv:2108.07732},
  year={2021}
}

@inproceedings{shidept,
title={De{PT}: Decomposed Prompt Tuning for Parameter-Efficient Fine-tuning},
author={Zhengxiang Shi and Aldo Lipani},
booktitle={The Twelfth International Conference on Learning Representations},
year={2024},
url={https://openreview.net/forum?id=KjegfPGRde}
}

@inproceedings{promptheory,
title={When Do Prompting and Prefix-Tuning Work? A Theory of Capabilities and Limitations},
author={Aleksandar Petrov and Philip Torr and Adel Bibi},
booktitle={The Twelfth International Conference on Learning Representations},
year={2024},
url={https://openreview.net/forum?id=JewzobRhay}
}

@inproceedings{nijkamp2023codegen,
title={CodeGen: An Open Large Language Model for Code with Multi-Turn Program Synthesis},
author={Erik Nijkamp and Bo Pang and Hiroaki Hayashi and Lifu Tu and Huan Wang and Yingbo Zhou and Silvio Savarese and Caiming Xiong},
booktitle={The Eleventh International Conference on Learning Representations },
year={2023},
url={https://openreview.net/forum?id=iaYcJKpY2B_}
}

@article{llama3,
  title={The llama 3 herd of models},
  author={Dubey, Abhimanyu and Jauhri, Abhinav and Pandey, Abhinav and Kadian, Abhishek and Al-Dahle, Ahmad and Letman, Aiesha and Mathur, Akhil and Schelten, Alan and Yang, Amy and Fan, Angela and others},
  journal={arXiv preprint arXiv:2407.21783},
  year={2024}
}

@inproceedings{
promptback,
title={Prompt Tuning Strikes Back: Customizing Foundation Models with Low-Rank Prompt Adaptation},
author={Abhinav Jain and Swarat Chaudhuri and Thomas Reps and Chris Jermaine},
booktitle={The Thirty-eighth Annual Conference on Neural Information Processing Systems},
year={2024},
url={https://openreview.net/forum?id=SyMhGilvCv}
}

@inproceedings{xiao-etal-2023-decomposed,
    title = "Decomposed Prompt Tuning via Low-Rank Reparameterization",
    author = "Xiao, Yao  and
      Xu, Lu  and
      Li, Jiaxi  and
      Lu, Wei  and
      Li, Xiaoli",
    editor = "Bouamor, Houda  and
      Pino, Juan  and
      Bali, Kalika",
    booktitle = "Findings of the Association for Computational Linguistics: EMNLP 2023",
    month = dec,
    year = "2023",
    address = "Singapore",
    publisher = "Association for Computational Linguistics",
    url = "https://aclanthology.org/2023.findings-emnlp.890",
    doi = "10.18653/v1/2023.findings-emnlp.890",
    pages = "13335--13347",
}

@inproceedings{ma-etal-2022-xprompt,
    title = "{XP}rompt: Exploring the Extreme of Prompt Tuning",
    author = "Ma, Fang  and
      Zhang, Chen  and
      Ren, Lei  and
      Wang, Jingang  and
      Wang, Qifan  and
      Wu, Wei  and
      Quan, Xiaojun  and
      Song, Dawei",
    editor = "Goldberg, Yoav  and
      Kozareva, Zornitsa  and
      Zhang, Yue",
    booktitle = "Proceedings of the 2022 Conference on Empirical Methods in Natural Language Processing",
    month = dec,
    year = "2022",
    address = "Abu Dhabi, United Arab Emirates",
    publisher = "Association for Computational Linguistics",
    url = "https://aclanthology.org/2022.emnlp-main.758",
    doi = "10.18653/v1/2022.emnlp-main.758",
    pages = "11033--11047",
}

@inproceedings{rajpurkar-etal-2016-squad,
    title = "{SQ}u{AD}: 100,000+ Questions for Machine Comprehension of Text",
    author = "Rajpurkar, Pranav  and
      Zhang, Jian  and
      Lopyrev, Konstantin  and
      Liang, Percy",
    booktitle = "Proceedings of the 2016 Conference on Empirical Methods in Natural Language Processing",
    month = nov,
    year = "2016",
    address = "Austin, Texas",
    publisher = "Association for Computational Linguistics",
    url = "https://aclanthology.org/D16-1264",
    doi = "10.18653/v1/D16-1264",
    pages = "2383--2392",
}

@article{zhang2018record,
  title={Record: Bridging the gap between human and machine commonsense reading comprehension},
  author={Zhang, Sheng and Liu, Xiaodong and Liu, Jingjing and Gao, Jianfeng and Duh, Kevin and Van Durme, Benjamin},
  journal={arXiv preprint arXiv:1810.12885},
  year={2018},
url={https://arxiv.org/abs/1810.12885}
}

@inproceedings{ruckle-etal-2021-adapterdrop,
    title = "{AdapterDrop}: {O}n the Efficiency of Adapters in Transformers",
    author = {R{\"u}ckl{\'e}, Andreas  and
      Geigle, Gregor  and
      Glockner, Max  and
      Beck, Tilman  and
      Pfeiffer, Jonas  and
      Reimers, Nils  and
      Gurevych, Iryna},
    booktitle = "Proceedings of the 2021 Conference on Empirical Methods in Natural Language Processing",
    month = nov,
    year = "2021",
    publisher = "Association for Computational Linguistics",
    url = "https://aclanthology.org/2021.emnlp-main.626",
}

@article{touvron2023llama,
  title={Llama 2: Open foundation and fine-tuned chat models},
  author={Touvron, Hugo and Martin, Louis and Stone, Kevin and Albert, Peter and Almahairi, Amjad and Babaei, Yasmine and Bashlykov, Nikolay and Batra, Soumya and Bhargava, Prajjwal and Bhosale, Shruti and others},
  journal={arXiv preprint arXiv:2307.09288},
  year={2023},
  url={https://arxiv.org/abs/2307.09288}
}

@inproceedings{mahabadi2021parameter,
    title = "Parameter-efficient Multi-task Fine-tuning for Transformers via Shared Hypernetworks",
    author = "Karimi Mahabadi, Rabeeh  and
      Ruder, Sebastian  and
      Dehghani, Mostafa  and
      Henderson, James",
    booktitle = "Proceedings of the 59th Annual Meeting of the Association for Computational Linguistics and the 11th International Joint Conference on Natural Language Processing (Volume 1: Long Papers)",
    month = aug,
    year = "2021",
    address = "Online",
    publisher = "Association for Computational Linguistics",
    url = "https://aclanthology.org/2021.acl-long.47",
    doi = "10.18653/v1/2021.acl-long.47",
    pages = "565--576",
}

@inproceedings{ivison-peters-2022-hyperdecoders,
    title = "Hyperdecoders: Instance-specific decoders for multi-task {NLP}",
    author = "Ivison, Hamish  and
      Peters, Matthew",
    booktitle = "Findings of the Association for Computational Linguistics: EMNLP 2022",
    month = dec,
    year = "2022",
    address = "Abu Dhabi, United Arab Emirates",
    publisher = "Association for Computational Linguistics",
    url = "https://aclanthology.org/2022.findings-emnlp.124",
    pages = "1715--1730",
}

@inproceedings{asai-etal-2022-attempt,
    title = "{ATTEMPT}: Parameter-Efficient Multi-task Tuning via Attentional Mixtures of Soft Prompts",
    author = "Asai, Akari  and
      Salehi, Mohammadreza  and
      Peters, Matthew  and
      Hajishirzi, Hannaneh",
    booktitle = "Proceedings of the 2022 Conference on Empirical Methods in Natural Language Processing",
    month = dec,
    year = "2022",
    address = "Abu Dhabi, United Arab Emirates",
    publisher = "Association for Computational Linguistics",
    url = "https://aclanthology.org/2022.emnlp-main.446",
    pages = "6655--6672",
}

@inproceedings{ResidualPrompt,
    title = "Residual Prompt Tuning: improving prompt tuning with residual reparameterization",
    author = "Razdaibiedina, Anastasiia  and
      Mao, Yuning  and
      Khabsa, Madian  and
      Lewis, Mike  and
      Hou, Rui  and
      Ba, Jimmy  and
      Almahairi, Amjad",
    editor = "Rogers, Anna  and
      Boyd-Graber, Jordan  and
      Okazaki, Naoaki",
    booktitle = "Findings of the Association for Computational Linguistics: ACL 2023",
    month = jul,
    year = "2023",
    address = "Toronto, Canada",
    publisher = "Association for Computational Linguistics",
    url = "https://aclanthology.org/2023.findings-acl.421",
    doi = "10.18653/v1/2023.findings-acl.421",
    pages = "6740--6757",
}

@inproceedings{vu-etal-2022-spot,
    title = "{SP}o{T}: Better Frozen Model Adaptation through Soft Prompt Transfer",
    author = "Vu, Tu  and
      Lester, Brian  and
      Constant, Noah  and
      Al-Rfou{'}, Rami  and
      Cer, Daniel",
    booktitle = "Proceedings of the 60th Annual Meeting of the Association for Computational Linguistics (Volume 1: Long Papers)",
    month = may,
    year = "2022",
    address = "Dublin, Ireland",
    publisher = "Association for Computational Linguistics",
    url = "https://aclanthology.org/2022.acl-long.346",
    doi = "10.18653/v1/2022.acl-long.346",
    pages = "5039--5059",
}

@inproceedings{li-liang-2021-prefix,
    title = "Prefix-Tuning: Optimizing Continuous Prompts for Generation",
    author = "Li, Xiang Lisa  and
      Liang, Percy",
    booktitle = "Proceedings of the 59th Annual Meeting of the Association for Computational Linguistics and the 11th International Joint Conference on Natural Language Processing (Volume 1: Long Papers)",
    month = aug,
    year = "2021",
    address = "Online",
    publisher = "Association for Computational Linguistics",
    url = "https://aclanthology.org/2021.acl-long.353",
    doi = "10.18653/v1/2021.acl-long.353",
    pages = "4582--4597",
}

@inproceedings{lester-etal-2021-power,
    title = "The Power of Scale for Parameter-Efficient Prompt Tuning",
    author = "Lester, Brian  and
      Al-Rfou, Rami  and
      Constant, Noah",
    booktitle = "Proceedings of the 2021 Conference on Empirical Methods in Natural Language Processing",
    month = nov,
    year = "2021",
    address = "Online and Punta Cana, Dominican Republic",
    publisher = "Association for Computational Linguistics",
    url = "https://aclanthology.org/2021.emnlp-main.243",
    doi = "10.18653/v1/2021.emnlp-main.243",
    pages = "3045--3059",
}

@inproceedings{zhang-etal-2019-paws,
    title = "{PAWS}: Paraphrase Adversaries from Word Scrambling",
    author = "Zhang, Yuan  and
      Baldridge, Jason  and
      He, Luheng",
    booktitle = "Proceedings of the 2019 Conference of the North {A}merican Chapter of the Association for Computational Linguistics: Human Language Technologies, Volume 1 (Long and Short Papers)",
    month = jun,
    year = "2019",
    address = "Minneapolis, Minnesota",
    publisher = "Association for Computational Linguistics",
    url = "https://aclanthology.org/N19-1131",
    doi = "10.18653/v1/N19-1131",
    pages = "1298--1308",
}

@article{Khot_Sabharwal_Clark_2018, 
title={SciTaiL: A Textual Entailment Dataset from Science Question Answering}, volume={32}, url={https://ojs.aaai.org/index.php/AAAI/article/view/12022}, DOI={10.1609/aaai.v32i1.12022}, 
number={1}, journal={Proceedings of the AAAI Conference on Artificial Intelligence}, author={Khot, Tushar and Sabharwal, Ashish and Clark, Peter}, year={2018}, month={Apr.} }

@inproceedings{10.5555/2969239.2969312,
author = {Zhang, Xiang and Zhao, Junbo and LeCun, Yann},
title = {Character-Level Convolutional Networks for Text Classification},
year = {2015},
publisher = {MIT Press},
address = {Cambridge, MA, USA},
booktitle = {Proceedings of the 28th International Conference on Neural Information Processing Systems - Volume 1},
pages = {649–657},
numpages = {9},
location = {Montreal, Canada},
url={https://proceedings.neurips.cc/paper/2015/file/250cf8b51c773f3f8dc8b4be867a9a02-Paper.pdf},
series = {NIPS'15}
}

@inproceedings{ouyang2022training,
title={Training language models to follow instructions with human feedback},
author={Long Ouyang and Jeffrey Wu and Xu Jiang and Diogo Almeida and Carroll Wainwright and Pamela Mishkin and Chong Zhang and Sandhini Agarwal and Katarina Slama and Alex Gray and John Schulman and Jacob Hilton and Fraser Kelton and Luke Miller and Maddie Simens and Amanda Askell and Peter Welinder and Paul Christiano and Jan Leike and Ryan Lowe},
booktitle={Advances in Neural Information Processing Systems},
editor={Alice H. Oh and Alekh Agarwal and Danielle Belgrave and Kyunghyun Cho},
year={2022},
url={https://openreview.net/forum?id=TG8KACxEON}
}

@inproceedings{NEURIPS2022_0cde695b,
 author = {Liu, Haokun and Tam, Derek and Muqeeth, Mohammed and Mohta, Jay and Huang, Tenghao and Bansal, Mohit and Raffel, Colin A},
 booktitle = {Advances in Neural Information Processing Systems},
 editor = {S. Koyejo and S. Mohamed and A. Agarwal and D. Belgrave and K. Cho and A. Oh},
 pages = {1950--1965},
 publisher = {Curran Associates, Inc.},
 title = {Few-Shot Parameter-Efficient Fine-Tuning is Better and Cheaper than In-Context Learning},
 url = {https://proceedings.neurips.cc/paper_files/paper/2022/file/0cde695b83bd186c1fd456302888454c-Paper-Conference.pdf},
 volume = {35},
 year = {2022}
}

@inproceedings{wang-etal-2018-glue,
    title = "{GLUE}: A Multi-Task Benchmark and Analysis Platform for Natural Language Understanding",
    author = "Wang, Alex  and
      Singh, Amanpreet  and
      Michael, Julian  and
      Hill, Felix  and
      Levy, Omer  and
      Bowman, Samuel",
    booktitle = "Proceedings of the 2018 {EMNLP} Workshop {B}lackbox{NLP}: Analyzing and Interpreting Neural Networks for {NLP}",
    month = nov,
    year = "2018",
    address = "Brussels, Belgium",
    publisher = "Association for Computational Linguistics",
    url = "https://aclanthology.org/W18-5446",
    doi = "10.18653/v1/W18-5446",
    pages = "353--355",
}

@article{liu2021gpt,
    title={GPT Understands, Too},
    author={Liu, Xiao and Zheng, Yanan and Du, Zhengxiao and Ding, Ming and Qian, Yujie and Yang, Zhilin and Tang, Jie},
    journal={arXiv:2103.10385},
    year={2021},
    url={https://arxiv.org/abs/2103.10385}
}

@article{warstadt2019neural_cola,
    title = "Neural Network Acceptability Judgments",
    author = "Warstadt, Alex  and
      Singh, Amanpreet  and
      Bowman, Samuel R.",
    journal = "Transactions of the Association for Computational Linguistics",
    volume = "7",
    year = "2019",
    address = "Cambridge, MA",
    publisher = "MIT Press",
    url = "https://aclanthology.org/Q19-1040",
    doi = "10.1162/tacl_a_00290",
    pages = "625--641",
}

@inproceedings{socher-etal-2013-recursive,
    title = "Recursive Deep Models for Semantic Compositionality Over a Sentiment Treebank",
    author = "Socher, Richard  and
      Perelygin, Alex  and
      Wu, Jean  and
      Chuang, Jason  and
      Manning, Christopher D.  and
      Ng, Andrew  and
      Potts, Christopher",
    booktitle = "Proceedings of the 2013 Conference on Empirical Methods in Natural Language Processing",
    month = oct,
    year = "2013",
    address = "Seattle, Washington, USA",
    publisher = "Association for Computational Linguistics",
    url = "https://aclanthology.org/D13-1170",
    pages = "1631--1642",
}

@inproceedings{dolan-brockett-2005-automatically,
    title = "Automatically Constructing a Corpus of Sentential Paraphrases",
    author = "Dolan, William B.  and
      Brockett, Chris",
    booktitle = "Proceedings of the Third International Workshop on Paraphrasing ({IWP}2005)",
    year = "2005",
    url = "https://aclanthology.org/I05-5002",
}

@inproceedings{cer-etal-2017-semeval,
    title = "{S}em{E}val-2017 Task 1: Semantic Textual Similarity Multilingual and Crosslingual Focused Evaluation",
    author = "Cer, Daniel  and
      Diab, Mona  and
      Agirre, Eneko  and
      Lopez-Gazpio, I{\~n}igo  and
      Specia, Lucia",
    booktitle = "Proceedings of the 11th International Workshop on Semantic Evaluation ({S}em{E}val-2017)",
    month = aug,
    year = "2017",
    address = "Vancouver, Canada",
    publisher = "Association for Computational Linguistics",
    url = "https://aclanthology.org/S17-2001",
    doi = "10.18653/v1/S17-2001",
    pages = "1--14",
}

@inproceedings{williams-etal-2018-broad,
    title = "A Broad-Coverage Challenge Corpus for Sentence Understanding through Inference",
    author = "Williams, Adina  and
      Nangia, Nikita  and
      Bowman, Samuel",
    booktitle = "Proceedings of the 2018 Conference of the North {A}merican Chapter of the Association for Computational Linguistics: Human Language Technologies, Volume 1 (Long Papers)",
    month = jun,
    year = "2018",
    address = "New Orleans, Louisiana",
    publisher = "Association for Computational Linguistics",
    url = "https://aclanthology.org/N18-1101",
    pages = "1112--1122",
}

@inproceedings{levesque2012winograd_wnli,
  title={The winograd schema challenge},
  author={Levesque, Hector and Davis, Ernest and Morgenstern, Leora},
  booktitle={Thirteenth International Conference on the Principles of Knowledge Representation and Reasoning},
  year={2012},
  url={https://cdn.aaai.org/ocs/4492/4492-21843-1-PB.pdf}
}

@article{10.1145/3474381,
author = {Sakaguchi, Keisuke and Bras, Ronan Le and Bhagavatula, Chandra and Choi, Yejin},
title = {WinoGrande: An Adversarial Winograd Schema Challenge at Scale},
year = {2021},
issue_date = {September 2021},
publisher = {Association for Computing Machinery},
address = {New York, NY, USA},
volume = {64},
number = {9},
issn = {0001-0782},
url = {https://doi.org/10.1145/3474381},
doi = {10.1145/3474381},
journal = {Commun. ACM},
month = {aug},
pages = {99–106},
numpages = {8}
}

@inproceedings{fisch-etal-2019-mrqa,
    title = "{MRQA} 2019 Shared Task: Evaluating Generalization in Reading Comprehension",
    author = "Fisch, Adam  and
      Talmor, Alon  and
      Jia, Robin  and
      Seo, Minjoon  and
      Choi, Eunsol  and
      Chen, Danqi",
    booktitle = "Proceedings of the 2nd Workshop on Machine Reading for Question Answering",
    month = nov,
    year = "2019",
    address = "Hong Kong, China",
    publisher = "Association for Computational Linguistics",
    url = "https://aclanthology.org/D19-5801",
    doi = "10.18653/v1/D19-5801",
    pages = "1--13",
}

@inproceedings{houlsby2019parameter,
  title={Parameter-efficient transfer learning for NLP},
  author={Houlsby, Neil and Giurgiu, Andrei and Jastrzebski, Stanislaw and Morrone, Bruna and De Laroussilhe, Quentin and Gesmundo, Andrea and Attariyan, Mona and Gelly, Sylvain},
  booktitle={International Conference on Machine Learning},
  pages={2790--2799},
  year={2019},
  url={http://proceedings.mlr.press/v97/houlsby19a/houlsby19a.pdf}
}

@inproceedings{hu2021lora,
  title={LoRA: Low-Rank Adaptation of Large Language Models},
  author={Hu, Edward J and Wallis, Phillip and Allen-Zhu, Zeyuan and Li, Yuanzhi and Wang, Shean and Wang, Lu and Chen, Weizhu and others},
  booktitle={International Conference on Learning Representations},
  year={2021},
  url={https://openreview.net/forum?id=nZeVKeeFYf9}
}

@inproceedings{
karimi2021compacter,
title={Compacter: Efficient Low-Rank Hypercomplex Adapter Layers},
author={Rabeeh Karimi Mahabadi and James Henderson and Sebastian Ruder},
booktitle={Advances in Neural Information Processing Systems},
editor={A. Beygelzimer and Y. Dauphin and P. Liang and J. Wortman Vaughan},
year={2021},
url={https://openreview.net/forum?id=bqGK5PyI6-N}
}

@inproceedings{zaken2021bitfit,
    title = "{B}it{F}it: Simple Parameter-efficient Fine-tuning for Transformer-based Masked Language-models",
    author = "Ben Zaken, Elad  and
      Goldberg, Yoav  and
      Ravfogel, Shauli",
    booktitle = "Proceedings of the 60th Annual Meeting of the Association for Computational Linguistics (Volume 2: Short Papers)",
    month = may,
    year = "2022",
    address = "Dublin, Ireland",
    publisher = "Association for Computational Linguistics",
    url = "https://aclanthology.org/2022.acl-short.1",
    doi = "10.18653/v1/2022.acl-short.1",
    pages = "1--9",
}

@inproceedings{guo2021diff,
    title = "Parameter-Efficient Transfer Learning with Diff Pruning",
    author = "Guo, Demi  and
      Rush, Alexander  and
      Kim, Yoon",
    booktitle = "Proceedings of the 59th Annual Meeting of the Association for Computational Linguistics and the 11th International Joint Conference on Natural Language Processing (Volume 1: Long Papers)",
    month = aug,
    year = "2021",
    address = "Online",
    publisher = "Association for Computational Linguistics",
    url = "https://aclanthology.org/2021.acl-long.378",
    doi = "10.18653/v1/2021.acl-long.378",
    pages = "4884--4896",
}

@inproceedings{he2022hyperprompt,
  title={Hyperprompt: Prompt-based task-conditioning of transformers},
  author={He, Yun and Zheng, Steven and Tay, Yi and Gupta, Jai and Du, Yu and Aribandi, Vamsi and Zhao, Zhe and Li, YaGuang and Chen, Zhao and Metzler, Donald and others},
  booktitle={International Conference on Machine Learning},
  pages={8678--8690},
  year={2022},
  url={https://proceedings.mlr.press/v162/he22f/he22f.pdf}
}

@inproceedings{
sunglst,
title={{LST}: Ladder Side-Tuning for Parameter and Memory Efficient Transfer Learning},
author={Yi-Lin Sung and Jaemin Cho and Mohit Bansal},
booktitle={Advances in Neural Information Processing Systems},
editor={Alice H. Oh and Alekh Agarwal and Danielle Belgrave and Kyunghyun Cho},
year={2022},
url={https://openreview.net/forum?id=isPnnaTZaP5}
}

@article{t5,
author = {Raffel, Colin and Shazeer, Noam and Roberts, Adam and Lee, Katherine and Narang, Sharan and Matena, Michael and Zhou, Yanqi and Li, Wei and Liu, Peter J.},
title = {Exploring the Limits of Transfer Learning with a Unified Text-to-Text Transformer},
year = {2020},
issue_date = {January 2020},
publisher = {JMLR.org},
volume = {21},
number = {1},
issn = {1532-4435},
journal = {J. Mach. Learn. Res.},
month = {jan},
articleno = {140},
numpages = {67},
url={https://dl.acm.org/doi/abs/10.5555/3455716.3455856}
}

@inproceedings{MPT,
title={Multitask Prompt Tuning Enables Parameter-Efficient Transfer Learning},
author={Zhen Wang and Rameswar Panda and Leonid Karlinsky and Rogerio Feris and Huan Sun and Yoon Kim},
booktitle={The Eleventh International Conference on Learning Representations },
year={2023},
url={https://openreview.net/forum?id=Nk2pDtuhTq}
}

@inproceedings{khashabi-etal-2018-looking,
    title = "Looking Beyond the Surface: A Challenge Set for Reading Comprehension over Multiple Sentences",
    author = "Khashabi, Daniel  and
      Chaturvedi, Snigdha  and
      Roth, Michael  and
      Upadhyay, Shyam  and
      Roth, Dan",
    booktitle = "Proceedings of the 2018 Conference of the North {A}merican Chapter of the Association for Computational Linguistics: Human Language Technologies, Volume 1 (Long Papers)",
    month = jun,
    year = "2018",
    address = "New Orleans, Louisiana",
    publisher = "Association for Computational Linguistics",
    url = "https://aclanthology.org/N18-1023",
    doi = "10.18653/v1/N18-1023",
    pages = "252--262",
}

@inproceedings{clark-etal-2019-boolq,
    title = "{B}ool{Q}: Exploring the Surprising Difficulty of Natural Yes/No Questions",
    author = "Clark, Christopher  and
      Lee, Kenton  and
      Chang, Ming-Wei  and
      Kwiatkowski, Tom  and
      Collins, Michael  and
      Toutanova, Kristina",
    booktitle = "Proceedings of the 2019 Conference of the North {A}merican Chapter of the Association for Computational Linguistics: Human Language Technologies, Volume 1 (Long and Short Papers)",
    month = jun,
    year = "2019",
    address = "Minneapolis, Minnesota",
    publisher = "Association for Computational Linguistics",
    url = "https://aclanthology.org/N19-1300",
    doi = "10.18653/v1/N19-1300",
    pages = "2924--2936",
}

@inproceedings{pilehvar-camacho-collados-2019-wic,
    title = "{W}i{C}: the Word-in-Context Dataset for Evaluating Context-Sensitive Meaning Representations",
    author = "Pilehvar, Mohammad Taher  and
      Camacho-Collados, Jose",
    booktitle = "Proceedings of the 2019 Conference of the North {A}merican Chapter of the Association for Computational Linguistics: Human Language Technologies, Volume 1 (Long and Short Papers)",
    month = jun,
    year = "2019",
    publisher = "Association for Computational Linguistics",
    url = "https://aclanthology.org/N19-1128",
}

@inproceedings{de2019commitmentbank,
  title={The commitmentbank: Investigating projection in naturally occurring discourse},
  author={De Marneffe, Marie-Catherine and Simons, Mandy and Tonhauser, Judith},
  booktitle={proceedings of Sinn und Bedeutung},
  volume={23},
  pages={107--124},
  year={2019},
  url={https://semanticsarchive.net/Archive/Tg3ZGI2M/Marneffe.pdf}
}

@inproceedings{wang2019superglue,
  author       = {Alex Wang and
                  Yada Pruksachatkun and
                  Nikita Nangia and
                  Amanpreet Singh and
                  Julian Michael and
                  Felix Hill and
                  Omer Levy and
                  Samuel R. Bowman},
  title        = {SuperGLUE: {A} Stickier Benchmark for General-Purpose Language Understanding
                  Systems},
  year         = {2019},
booktitle={arxiv},
  url          = {http://arxiv.org/abs/1905.00537},
}

@inproceedings{giampiccolo-etal-2007-third,
    title = "The Third {PASCAL} Recognizing Textual Entailment Challenge",
    author = "Giampiccolo, Danilo  and
      Magnini, Bernardo  and
      Dagan, Ido  and
      Dolan, Bill",
    booktitle = "Proceedings of the {ACL}-{PASCAL} Workshop on Textual Entailment and Paraphrasing",
    month = jun,
    year = "2007",
    address = "Prague",
    publisher = "Association for Computational Linguistics",
    url = "https://aclanthology.org/W07-1401",
    pages = "1--9",
}

@inproceedings{yang-etal-2018-hotpotqa,
    title = "{H}otpot{QA}: A Dataset for Diverse, Explainable Multi-hop Question Answering",
    author = "Yang, Zhilin  and
      Qi, Peng  and
      Zhang, Saizheng  and
      Bengio, Yoshua  and
      Cohen, William  and
      Salakhutdinov, Ruslan  and
      Manning, Christopher D.",
    booktitle = "Proceedings of the 2018 Conference on EMNLP",
    month = oct # "-" # nov,
    year = "2018",
    address = "Brussels, Belgium",
    publisher = "Association for Computational Linguistics",
    url = "https://aclanthology.org/D18-1259",
}

@article{dunn2017searchqa,
  title={Searchqa: A new q\&a dataset augmented with context from a search engine},
  author={Dunn, Matthew and Sagun, Levent and Higgins, Mike and Guney, V Ugur and Cirik, Volkan and Cho, Kyunghyun},
  journal={arXiv preprint arXiv:1704.05179},
  year={2017},
  url={https://arxiv.org/abs/1704.05179}
}

@inproceedings{trischler-etal-2017-newsqa,
    title = "{N}ews{QA}: A Machine Comprehension Dataset",
    author = "Trischler, Adam  and
      Wang, Tong  and
      Yuan, Xingdi  and
      Harris, Justin  and
      Sordoni, Alessandro  and
      Bachman, Philip  and
      Suleman, Kaheer",
    booktitle = "Proceedings of the 2nd Workshop on Representation Learning for {NLP}",
    month = aug,
    year = "2017",
    address = "Vancouver, Canada",
    publisher = "Association for Computational Linguistics",
    url = "https://aclanthology.org/W17-2623",
    doi = "10.18653/v1/W17-2623",
    pages = "191--200",
}

@article{kwiatkowski-etal-2019-natural,
    title = "Natural Questions: A Benchmark for Question Answering Research",
    author = "Kwiatkowski, Tom  and
      Palomaki, Jennimaria  and
      Redfield, Olivia  and
      Collins, Michael  and
      Parikh, Ankur  and
      Alberti, Chris  and
      Epstein, Danielle  and
      Polosukhin, Illia  and
      Devlin, Jacob  and
      Lee, Kenton  and
      Toutanova, Kristina  and
      Jones, Llion  and
      Kelcey, Matthew  and
      Chang, Ming-Wei  and
      Dai, Andrew M.  and
      Uszkoreit, Jakob  and
      Le, Quoc  and
      Petrov, Slav",
    journal = "Transactions of the Association for Computational Linguistics",
    volume = "7",
    year = "2019",
    address = "Cambridge, MA",
    publisher = "MIT Press",
    url = "https://aclanthology.org/Q19-1026",
    doi = "10.1162/tacl_a_00276",
    pages = "452--466",
}

@inproceedings{kopiczko2024vera,
title={Ve{RA}: Vector-based Random Matrix Adaptation},
author={Dawid Jan Kopiczko and Tijmen Blankevoort and Yuki M Asano},
booktitle={The Twelfth International Conference on Learning Representations},
year={2024},
url={https://openreview.net/forum?id=NjNfLdxr3A}
}

@misc{gsm8k,
      title={Training Verifiers to Solve Math Word Problems}, 
      author={Karl Cobbe and Vineet Kosaraju and Mohammad Bavarian and Mark Chen and Heewoo Jun and Lukasz Kaiser and Matthias Plappert and Jerry Tworek and Jacob Hilton and Reiichiro Nakano and Christopher Hesse and John Schulman},
      year={2021},
      eprint={2110.14168},
      archivePrefix={arXiv},
      primaryClass={cs.LG},
      url={https://arxiv.org/abs/2110.14168}, 
}

\newpage
\appendix


\section*{Appendix Overview}

The appendix is structured as follows:
\paragraph{Appendix \ref{sec:DePT_theory}} provides a theoretical analysis towards DePT.  We combine the theoretical analysis from Section \ref{theoretical analysis} and the Appendix \ref{sec:DePT_theory} to theoretically explain why ADePT performs better than DePT.
\paragraph{Appendix \ref{sec:additional_experiments}} provides additional experiments to further analyze our proposed ADePT.
\paragraph{Appendix \ref{sec:additional_implementation_details}} provides a more detailed implementation of our experiments.
\paragraph{Appendix \ref{ref:datasets1}} provides a detailed description of datasets.
\paragraph{Appendix \ref{sec:hyperparameters}} provides detailed hyperparameters of our experiments.

\section{How Decomposed Prompt Tuning Affect the first multi-head self-attention layer?}
\label{sec:DePT_theory}

In this section, we analyze how the DePT affects the first multi-head self-attention layer.

Given the soft prompt $\bm{P} = [\bm{p}_1, \bm{p}_2, \dots, \bm{p}_s] \in \mathbb{R}^{l \times d}$ and the offset embeddings $\Delta \bm{E} = \bm{A} \bm{B} \in \mathbb{R}^{s \times d}$, the output of a query vector $\bm{x}_i$ passing through the single-head self-attention $\mathcal{H}$ is formulated as,

\begin{align}
\begin{aligned}
    &\bm{o}_i^{\text{DePT}}   =  \text{Attention}\left(\left( \bm{e}_i + \Delta \bm{e}_i\right) \bm{W}_Q^\mathcal{H} , \text{concat}[\bm{P},\bm{E} + \Delta \bm{E} ] \bm{W}_K^\mathcal{H} , \text{concat}[\bm{P},\bm{E} + \Delta \bm{E} ] 
 \bm{W}_V^\mathcal{H} \right),\\
    & = \sum_{k=1}^{l} A_{ik} \bm{p}_k \bm{W}_V^{\mathcal{H}} + \left(1 - \sum_{k=1}^{l} A_{ik}\right) \text{Softmax}\left( \left(\left( \bm{e}_i + \Delta \bm{e}_i \right) \bm{W}_Q^{ \mathcal{H}} \right)\left(\left(\bm{E}+\Delta \bm{E}\right) \bm{W}_K^{\mathcal{H}}  \right)^T\right) \Delta \bm{E}  \bm{W}_V^{\mathcal{H}}  \\
    & + (1 - \sum_{k=1}^{l} A_{ik}) \text{Softmax}\left( \left(\left( \bm{e}_i + \Delta \bm{e}_i \right) \bm{W}_Q^{\mathcal{H}} \right)\left( \left(\bm{E}+\Delta \bm{E}\right) 
 \bm{W}_K^{\mathcal{H}} \right)^T\right) \bm{E} \bm{W}_V^{\mathcal{H}} ,\\
   & \text{with}\quad A_{ik} = \frac{\exp\left(\left(\left(\bm{e}_i + \Delta \bm{e}_i \right) \bm{W}_Q^{\mathcal{H}} \right)\left(\bm{p}_k \bm{W}_K^{\mathcal{H}}\right)^\top \right)}{C},\\
   & C = \sum_{k=1}^{l}\exp\left(\left(\left(\bm{e}_i + \Delta \bm{e}_i \right) \bm{W}_Q^\mathcal{H} \right) \left(\bm{p}_k \bm{W}_K^\mathcal{H}\right)^\top \right) \\
   & + \sum_{j=1}^{s} \exp\left( \left(\left(\bm{e}_i + \Delta \bm{e}_i \right) 
 \bm{W}_Q^{\mathcal{H}} \right) \left(\left(\bm{e}_j+\Delta \bm{e}_j\right) \bm{W}_K^\mathcal{H}\right)^\top \right),
    \label{eq:DePT_First_layer}
\end{aligned}
\end{align}
where $ A_{ik}$ is the attention score gives to the prefix vector $\bm{p}_k$ for a given query vector $\bm{e}_i$. We can observe that, in the first transformer layer, DePT can change the relative attention patterns. However, compared to ADePT, the attention patterns change along the change of position. Also, DePT cannot add a bias dependent on model input in the first transformer layer.

\clearpage
\section{ADDITIONAL EXPERIMENTS}
\label{sec:additional_experiments}
\subsection{Ablation Study}
\begin{table}[!h]
\centering
\caption{
The comparison of training time based on T5-3B model. ``h'' means hours.}

\vspace{0.1em}
\resizebox{0.45\columnwidth}{!}{
\begin{tabular}{lccc}
\toprule
\bf Method  & \bf\#Para              & \bf NQ   & \bf HP \\  
\midrule
LoRA & 25.8M& 9.73 h &  9.63 h\\
PT         &    153.6K    &  12.60 h &	12.53 h \\
DePT  &   153.6K   &  12.60 h &	 12.53 h  \\  
ADePT (ours) &  152.9K     &  12.67 h  & 12.58 h  \\  
\bottomrule
\end{tabular}
}
\label{table:training_time}
\vspace{-0.5em}
\end{table}

\begin{table}[!h]
\centering
\caption{
the experimental results of fine-tuning both the prompt and the embedding matrix, as well as our proposed ADePT based on the T5-base model for RTE.}

\vspace{0.1em}
\resizebox{0.55\columnwidth}{!}{
\begin{tabular}{lccc}
\toprule
\bf Method  & \bf\#Para              & \bf RTE \\  
\midrule
Finetuning prompt and embedding matrix & 24.8M  &  76.3\\  
ADePT (ours) & 76.1K     &  82.0   \\  
\bottomrule
\end{tabular}
}
\label{table:fine_emebdding}
\vspace{-0.5em}
\end{table}

\begin{table}[!h]
\centering
\caption{
the experimental results of ADePT/DePT when used without the token offsets but only learned soft prompt. The results are based on the T5-base model and the RTE task.}

\vspace{0.1em}
\resizebox{0.55\columnwidth}{!}{
\begin{tabular}{lccc}
\toprule
\bf Method  & \bf {With token offsets}              & \bf {Without token offsets}\\  
\midrule
DePT & 79.1  &  78.4\\  
ADePT (ours) & 82.0     &  58.3   \\  
\bottomrule
\end{tabular}
}
\label{table:without_token_embeddings}
\vspace{-0.5em}
\end{table}

\begin{table}[!h]
\centering
\caption{
The results of ADePT with different bottleneck sizes using the T5-base model on the RTE.}

\vspace{0.1em}
\resizebox{0.5\columnwidth}{!}{
\begin{tabular}{lcccc}
\toprule
\bf The size of the Bottleneck   & \bf 5             & \bf 10  & \bf 20 & \bf 30 \\  
\midrule

ADePT (ours) & 78.4 & 82.0    &  82.0  & 79.9  \\  
\bottomrule
\end{tabular}
}
\label{table:different_bottleneck_size_affect}
\vspace{-0.5em}
\end{table}

\begin{table}[!h]
\centering
\caption{
The results of ADePT with soft prompt lengths using the T5-base model on the RTE.}

\vspace{0.1em}
\resizebox{0.6\columnwidth}{!}{
\begin{tabular}{lcccccc}
\toprule
\bf The length of soft prompt   & \bf 20             & \bf 40  & \bf 50 &  \bf 60 & \bf 70 & \bf 80 \\  
\midrule

ADePT (ours) & 72.7 &  77.7 &  79.1 &  82.0  & 80.6 & 79.1  \\  
\bottomrule
\end{tabular}
}
\label{table:different_prompt_length_affect}
\vspace{-0.5em}
\end{table}

\begin{table}[!h]
\centering
\caption{
The comparison of ADePT with only feed-forward neural network and the original ADePT based on T5-base model for the RTE.}

\vspace{0.1em}
\resizebox{0.5\columnwidth}{!}{
\begin{tabular}{lccc}
\toprule
\bf Method  & \bf\#Para              & \bf RTE \\  
\midrule
Only Feed-forward Neural Network & 76.1K  &  73.4\\  
ADePT (ours) & 76.1K     &  82.0   \\  
\bottomrule
\end{tabular}
}
\label{table:all_FFN}
\vspace{-0.5em}
\end{table}

Table \ref{table:training_time} indicates that PT, DePT, and ADePT need similar training time. The PT-family method needs longer training time than LoRA due to the longer input sequence.

Table \ref{table:fine_emebdding} shows the experimental results of fine-tuning both the prompt and the embedding matrix, as well as our proposed ADePT. We use the same length soft prompt, and we search learning rates for prompt matrix from \{3e-1, 4e-1, 5e-1\} and embedding matrix from \{1e-3, 1e-4, 1e-5\}. We observe that fine-tuning both the prompt and the embedding matrix underperforms our proposed ADePT. Additionally, this method requires fine-tuning a large number of parameters, which may lead to overfitting.

Table \ref{table:without_token_embeddings} show the experimental results of ADePT/DePT when used without the token offsets but only learned soft prompt. We can observe that the token offsets of ADePT play a much more important role than DePT.

Table \ref{table:different_bottleneck_size_affect} presents how the size of the bottleneck affects the performance of the RTE task. We can observe that an overly small or overly large bottleneck size will cause a decline in performance. When it is too small, it can lead to under-fitting; when it is too large, it can lead to over-fitting.

Table \ref{table:different_prompt_length_affect} presents how the length of the prompt affects the performance on the RTE task. We can observe that performance drops significantly when the prompt length is less than 40. Performance is optimal when the prompt length is 50 or 60, but it decreases when the prompt length is too large, possibly due to overfitting.

Table \ref{table:all_FFN} shows that the performance on the RTE task when all parameters are relocated to the learnable projection (prompt length = 0, bottleneck size = 49, trainable parameters = 76.1k) is 73.4, indicating the soft prompt is necessary.

\subsection{Few-shot Learning}

\begin{table}[!h]
\centering

\caption{
Few-shot learning results, obtained from three runs, with $k$ = \{4, 16, 32\} training samples on the BooQ, CB and SciTail datasets. Baseline results are directly quoted from \citet{shidept}.
}
\vspace{0.1em}
\resizebox{0.8\columnwidth}{!}{

\begin{tabular}{lccccccccccc}
\toprule  

\multirow{2}{*}{\bf Task}&\bf  $k$-shot &\bf {Full Finetuning} &\bf AD &\bf PT &\bf ST &\bf HF &\bf  (IA)$^3$ & \bf ATP &\bf MPT   &   \bf DePT  & \bf ADePT (ours)\\
                         &\bf  \#Para   & 220M  & 1.9M  & 76.8K & 76.8K & 638K  &   55.3K     & 232K   & 77.6K   & 76.8K & 76.1K\\ \midrule
\multirow{3}{*}{BoolQ}
                         & 4        & 50.5  & 53.4  & 61.6  & 50.5  & 48.0 &  56.7 & 61.8   & 62.2 & 62.7$_{5.4}$ & 68.7$_{0.4}$\\       
                         & 16       & 56.5  & 51.4  & 61.9  & 50.6  & 50.2 &  62.0 & 60.0   & 63.3 & 66.9$_{4.4}$ & 69.9$_{1.3}$\\          
                         & 32       & 58.4  & 54.5  & 61.7  & 61.2  & 58.3 &  67.2 & 65.3   &  68.9 &  67.2$_{3.4}$ & 70.0$_{1.2}$ \\ \midrule  
\multirow{3}{*}{CB}
                         & 4        & 57.7  & 51.1  & 53.5  & 71.4  & 60.7 &  65.5 & 67.9   & 73.6 & 75.0$_{5.1}$ & 32.1$_{2.6}$ \\  
                         & 16       & 77.0  & 74.8  & 63.5  & 64.3  & 76.3 &  71.4 & 71.4   & 78.6 & 78.6$_{4.3}$ & 36.7$_{2.3}$\\  
                         & 32       & 80.0  & 74.8  & 67.8  & 64.3  & 81.4 &  75.0 & 78.5   & 82.1 & 82.1$_{2.3}$ & 39.5$_{3.1}$\\ \midrule  

\multirow{3}{*}{SciTail}
                         & 4        & 79.6  & 79.5  & 57.7  & 69.6  & 82.0 &  65.4 & 80.2   &  80.2 &  78.1$_{2.5}$ &  76.9$_{3.2}$\\     
                         & 16       & 80.0  & 83.2  & 60.8  & 71.9  & 86.5 &  74.4 & 79.5   &  87.3 &  78.5$_{1.4}$ & 82.1$_{2.0}$ \\   
                         & 32       & 81.9  & 85.0  & 60.2  & 71.9  & 85.8 &  80.4 & 80.2   &  86.3 &  85.4$_{3.1}$ &  82.6$_{2.6}$\\     
\bottomrule
\end{tabular}}
\label{table:few_Shot}
\vspace{-0.5em}
\end{table}

Table \ref{table:few_Shot} shows the few-shot learning results with $k$ = \{4, 16, 32\} training samples on BoolQ, CB and SciTail datasets. We pre-train both the soft prompt and the feed-forward neural network on source tasks and select the best checkpoint to initialize the parameters. We can observe that ADePT performs best on the BoolQ dataset, performs well on the SciTail dataset, and performs the worst on the CB dataset. This might indicate that ADePT is unsuitable for few-shot learning, which is reasonable since learning the embedding offsets for each token using a feed-forward neural network requires considerable training samples.

\subsection{Unsuccessful Attempts}
We attempted to perform instruction tuning on a larger scale of the GSM8K dataset \citep{gsm8k} using Llama3 8B, with 7,473 examples in the training set and 1,319 examples in the test set. The goal of this instruction tuning was to fine-tune the model to better solve mathematical reasoning problems in the GSM8K dataset. However, unfortunately, we were not successful in our training attempts, including both ADePT and DePT. This might be because the simple low-rank matrix multiplication and shallow neural networks are unable to fit the input token embedding offsets required by large datasets when working with large models that have a very large vocabulary. Addressing this issue will be a direction for our future research.

\clearpage
\section{Additional Implementation Details}

\label{sec:additional_implementation_details}
We implement our experiments by using Pytorch\footnote{\url{https://pytorch.org/}}, Huggingface Transformers\footnote{\url{https://github.com/huggingface/transformers}}, and Huggingface PEFT \footnote{\url{https://github.com/huggingface/peft}}. We evaluate our proposed ADePT in four PLMs, \textit{i.e.}, T5-base model\footnote{\url{https://huggingface.co/google-t5/t5-base}}, T5-3B model\footnote{\url{https://huggingface.co/google-t5/t5-3b}}, CodeGen-350M\footnote{\url{ https://huggingface.co/Salesforce/codegen-350M-mono}} and Llama3-8B\footnote{\url{https://huggingface.co/meta-llama/Meta-Llama-3-8B}}. Following \citet{asai-etal-2022-attempt,MPT,shidept}, we train the T5 model using the original checkpoint rather than the LM-adapted 1.1 version \citep{lester-etal-2021-power}. For the T5-3B model, due to the limitations of computational resources, we select the several most convincing datasets to evaluate our proposed ADePT. We think that convincing datasets should have a large training dataset and large test dataset, and be challenging for the  T5-base model. We use the criteria of more than 70,000 training samples, an accuracy/F1 score of less than 90\% on the T5-base model, and more than 4,000 test samples to select the datasets to test our proposed ADePT on the T5-3B model. \citet{shidept} found that training PT for additional steps typically leads to performance improvements, and we follow this setting to train our proposed ADePT. We measure the latency of ADePT by running a feed-forward neural network for each token in real time.

For the small datasets ($<70,000$ training samples), following \citet{shidept}, we search the learning rates for the soft prompt from \{$3e-1, 4e-1, 5e-1$\} and for the feed-forward neural network from \{$1e-4, 1e-5$\}. We also search for the prompt length from \{$20, 40, 60, 80$\} with corresponding bottleneck sizes of \{$39, 29, 19, 9$\} to ensure the number of trainable parameters remains below 76.8K.
For the large datasets ($>70,000$ training samples), we set the prompt length as $60$, the bottleneck size as $19$, the prompt learning rate as $3e-1$, and the feed-forward neural network learning rate as $1e-4$. For decoder-only PLMs, following \citet{promptback}, we use $10$ virtual tokens for PT, $7$ virtual tokens and rank $r_s = 3$ for DePT, $7$ virtual tokens and bottleneck size $r=1$ for ADePT, and rank $16$ for LoRA. For the MBPP benchmark, following \citet{promptback}, we use learning rates of $1e-3$ for the prompting-style tuning method and $1e-4$ for LoRA.

Following \citet{shidept}, for the T5-base model and the small datasets, we train the model for 30,000 steps; for the T5-base model and the large datasets, we train the model for 300,000 steps. For the T5-3B model, we train the model for 30,000 steps.
In each trial of the t5-base model and the T5-3B model, we evaluate the performance every 1,000 steps and select the best checkpoint based on the optimal performance on the evaluation set. For the MBPP benchmarks, following \citet{promptback}, we train the model for 10 epochs. We train the model with a batch size of 32, except for the MBPP benchmark, where we use a batch size of 4. We typically use a maximum sequence length of 256, except for SuperGLUE-MultiRC, where the maximum sequence length is 348, and MRQA, where it is 512.

For the few-shot learning, following the prior works \citep{asai-etal-2022-attempt,MPT,shidept}, we first pre-train five source tasks (\textit{i.e.}, MNLI, QQP, SST-2, SQUAD, and ReCoRD), and then select the best checkpoint to use as the initialization for few-shot training.

%

\clearpage
\section{Datasets}
\label{ref:datasets1}

\begin{table*}[!h]
\begin{center}
\centering
\caption{The datasets assessed in this study are described as follows. The term ``Source Length'' refers to the average length of source sentences in the training set, whereas ``Target Length'' indicates the average length of target sentences in the training set. The term ``Train'' refers to the number of samples in the training set, whereas ``Valid'' and ``Test'' indicate the number of samples in the validation set and test set, respectively. The term ``Type'' refers to the task type of the dataset.
}
\resizebox{\columnwidth}{!}{%
\begin{tabular}{lrrrrrl}
\toprule
\multicolumn{7}{c}{\bf \textit{GLUE Benchmark}} \\
\midrule
\bf Dataset & \bf Source Length & \bf Target Length & \bf Train & \bf Valid & \bf Test & \bf Type  \\
\midrule
MNLI  & 31.8 & 1.0  & 392,702 & 9,832 & 9,815  & NLI \\ 
QQP   & 24.1 & 1.0  & 362,846 & 1,000 & 40,431 & Paraphrase \\
QNLI  & 38.4 & 1.0  & 103,743 & 1,000 & 5,463  & NLI \\ 
SST-2 & 10.4 & 1.0     & 66,349  & 1,000 & 872    & Sentiment  \\ 
STS-B & 21.9 & 1.0  & 5,749 & 750 & 750  & Sent. Similarity \\ 
MRPC  & 45.9 & 1.0  & 3,668 & 204 & 204 & Paraphrase \\ 
RTE   & 54.4 & 1.0  & 2,490 & 138 & 139 & NLI \\ 
CoLA  & 8.7 & 1.0  & 8,551 & 521 & 522 & Acceptability \\ 
\midrule
\multicolumn{7}{c}{\bf \textit{SuperGLUE Benchmark}} \\
\midrule
\bf Dataset & \bf Source & \bf Target & \bf Train & \bf Valid & \bf Test & \bf Type \\
\midrule
MultiRC     & 286.1 & 1.0 & 27,243 & 2,424 & 2,424 & Question Answering \\ 
BoolQ       & 108.3 & 1.0 & 9,427  & 1,635 & 1,635 & Question Answering \\ 
WiC         & 18.4 & 1.0 & 5,428  & 319   & 319 & Word Sense Disambiguation \\ 
WSC         & 28.1 & 1.0 & 554    & 52    & 52 & Common Sense Reasoning \\ 
CB          & 64.6 & 1.0 & 250    & 28    & 28 & NLI \\ 
ReCoRD      & 210.7 & 1.5 & 137,484 & 1,370 & 15,176 & Common Sense Reasoning \\
\midrule
\multicolumn{7}{c}{\bf \textit{MRQA 2019 Shared Task}} \\
\midrule
\bf Dataset & \bf Source & \bf Target & \bf Train & \bf Valid & \bf Test & \bf Type \\
\midrule
NaturalQuestions & 242.7 & 4.5 & 103,071 & 1,000 & 12836 & Question Answering \\ 
HotpotQA         & 225.7 & 2.6 & 71,928  & 1,000 & 5,901 & Question Answering \\ 
SearchQA         & 942.8 & 2.0 & 116,384 & 1,000 & 16,980 & Question Answering \\ 
NewsQA           & 615.5 & 5.1 & 73,160  & 1,000 & 4,212 &  Question Answering \\ 
\midrule
\multicolumn{7}{c}{\bf \textit{Other Datasets}} \\
\midrule
\bf Dataset & \bf Source & \bf Target & \bf Train & \bf Valid & \bf Test & \bf Type \\
\midrule
WinoGrande   & 23.8 & 1.0 & 39,398  & 1,000 & 1,267  & Common Sense Reasoning \\ 
YelpPolarity & 134.0 & 1.0 & 100,000 & 1,000 & 38,000 & Sentiment \\ 
SciTail      & 30.8 & 1.0 & 23,596  & 652   & 652    & NLI \\ 
PAWS         & 44.7 & 1.0 & 4,9401  & 8,000 & 8,000  & Sent. Similarity \\ 
\bottomrule
\end{tabular}
}
\end{center}
\label{appendix:datasets}

\end{table*}
\clearpage
\section{Hyperparameters}
\label{sec:hyperparameters}

\begin{table*}[!ht]
    \centering
    \small
    \caption{Hyperparameters of small datasets for ADePT on T5-base model.} 
    \begin{tabular}{cc}
        \toprule
        \textbf{Hyperparameter} & \textbf{Assignment}  \\
        \midrule
        number of steps & 30,000 steps (evaluate every 1,000 steps)\\
        \midrule
        batch size & 32 \\
        \midrule
        maximum learning rate ($\alpha_1$) & 3e-1, 4e-1, 5e-1 \\
        \midrule
        maximum learning rate ($\alpha_2$) & 1e-4, 1e-5\\
        \midrule
        length of the soft prompt ($m$) & 20, 40, 60, 80 \\
        \midrule
        maximum sequence length & 256 \\
        \midrule
        learning rate optimizer & AdamW \\
        \midrule
        Adam epsilon & 1e-6 \\
        \midrule
        Adam beta weights & 0.9, 0.98\\
        \midrule
        learning rate scheduler & Warmup linear \\
        \midrule
        Weight decay & 0.01 \\
        \midrule
        Warmup steps & 500 \\
        \bottomrule
    \end{tabular}
    \label{table:pft_hyperparameters1}
\end{table*}

\begin{table*}[!ht]
    \centering
    \small
    \caption{Hyperparameters of large datasets for ADePT on T5-base model.} 
    \begin{tabular}{cc}
        \toprule
        \textbf{Hyperparameter} & \textbf{Assignment}  \\
        \midrule
        number of steps & 300,000 steps (evaluate every 1,000 steps)\\
        \midrule
        batch size & 16 \\
        \midrule
        gradient accumulation steps & 2 \\
        \midrule
        maximum learning rate ($\alpha_1$) & 3e-1\\
        \midrule
        maximum learning rate ($\alpha_2$) & 1e-4\\
        \midrule
        length of the soft prompt ($m$) & 60 \\
        \midrule
        maximum sequence length & 512 \\
        \midrule
        learning rate optimizer & AdamW \\
        \midrule
        Adam epsilon & 1e-6 \\
        \midrule
        Adam beta weights & 0.9, 0.98\\
        \midrule
        learning rate scheduler & Warmup linear \\
        \midrule
        Weight decay & 0.01 \\
        \midrule
        Warmup steps & 500 \\
        \bottomrule
    \end{tabular}
    \label{table:pft_hyperparameters2}
\end{table*}

\begin{table*}[!th]
    \centering
    \small
    \caption{Hyperparameters for ADePT on T5-3B model.} 
    \begin{tabular}{cc}
        \toprule
        \textbf{Hyperparameter} & \textbf{Assignment}  \\
        \midrule
        number of steps & 30,000 steps (evaluate every 1,000 steps)\\
        \midrule
        batch size & 16 \\
        \midrule
        gradient accumulation steps & 2 \\
        \midrule
        maximum learning rate ($\alpha_1$) & 3e-1\\
        \midrule
        maximum learning rate ($\alpha_2$) & 1e-4\\
        \midrule
        length of the soft prompt ($m$) & 60 \\
        \midrule
        maximum sequence length & 512 \\
        \midrule
        learning rate optimizer & AdamW \\
        \midrule
        Adam epsilon & 1e-6 \\
        \midrule
        Adam beta weights & 0.9, 0.98\\
        \midrule
        learning rate scheduler & Warmup linear \\
        \midrule
        Weight decay & 0.01 \\
        \midrule
        Warmup steps & 500 \\
        \bottomrule
    \end{tabular}
    \label{table:pft_hyperparameters3}
\end{table*}

\clearpage

\end{document}